\documentclass[10pt,twocolumn,letterpaper]{article}

\usepackage[8]{cvpr} 

\usepackage{graphicx}
\usepackage{amsmath}
\usepackage{amssymb}
\usepackage{booktabs}
\usepackage[dvipsnames]{xcolor}
\usepackage{multirow}

\usepackage[pagebackref,breaklinks,colorlinks]{hyperref}

\usepackage[capitalize]{cleveref}
\crefname{section}{Sec.}{Secs.}
\Crefname{section}{Section}{Sections}
\Crefname{table}{Table}{Tables}
\crefname{table}{Tab.}{Tabs.}

\def\cvprPaperID{1} 
\def\confName{Precognition}
\def\confYear{2023}

\begin{document}

\title{A unified model for continuous conditional video prediction}

\author{Xi Ye\\
Polytechnique Montreal\\
{\tt\small xi.ye@polymtl.ca}
\and
Guillaume-Alexandre Bilodeau\\
Polytechnique Montreal\\
{\tt\small gabilodeau@polymtl.ca}
}
\maketitle

\begin{abstract}
    Different conditional video prediction tasks, like video future frame prediction and video frame interpolation, are normally solved by task-related models even though they share many common underlying characteristics. Furthermore, almost all conditional video prediction models can only achieve discrete prediction. In this paper, we propose a unified model that addresses these two issues at the same time. We show that conditional video prediction can be formulated as a neural process, which maps input spatio-temporal coordinates to target pixel values given context spatio-temporal coordinates and context pixel values. Specifically, we feed the implicit neural representation of coordinates and context pixel features into a Transformer-based non-autoregressive conditional video prediction model. Our task-specific models outperform previous work for video future frame prediction and video interpolation on multiple datasets. Importantly, the model is able to interpolate or predict with an arbitrary high frame rate, i.e., continuous prediction. Our source code is available at
   \url{https://npvp.github.io}.
\end{abstract}

\section{Introduction}

\label{sec:intro}
The human ability to anticipate dynamic changes in a scene is remarkable, for instance, we can effortlessly anticipate the possible position of a car in the next few seconds or envision the most recent movement of a jogger. Conditional video prediction models are essential for creating human-like intelligent agents, which have numerous applications, such as autonomous driving, robotics, and more. In this paper, we focus on two closely related conditional video prediction tasks, video future frame prediction (VFP), which consists in predicting future frames given some past frames, and video frame interpolation (VFI) with long temporal gap. Even though VFP and VFPI share some similarities, they have been tackled by totally different methods for a long time. For example, most VFP models depend on Convolutional-LSTMs (ConvLSTMs) to predict the future frames autoregressively \cite{denton2018, kwon2019, chang2021, lee2021}, and they are incapable of performing video interpolation. Meanwhile, most VFI methods capture the motion between input frames by estimating optical flow \cite{jiang2018, niklaus2020} or local convolution kernels \cite{niklaus2017, niklaus2021}. None of these methods are able to solve the VFP problem.

 Therefore, we address the problem of unifying multiple conditional video prediction tasks to solve them with a single model. Our motivation to propose a unified model is that multi-task learning is a good regularization for a better representation learning \cite{goodfellow2016}. Thus, we believe that a unified model is beneficial for each individual task. Additionally, one common problem for almost all conditional video prediction models is that they can only generate video with a fixed frame rate, i.e., discrete prediction. However, the real world is continuous over the spatio-temporal space. Therefore, we also aim to develop a conditional video prediction model that is able to recover the underlying continuous signal of the real world given a discrete dataset, and thus enable many useful applications, e.g., generating videos with an arbitrary high frame rate, or generating a climate video with irregular time interval \cite{park2021b}.
 
 To address these two problems, we propose a novel unsupervised continuous conditional video prediction method based on neural processes (NPs) \cite{garnelo2018} and implicit neural representations (INRs) \cite{tancik2020, sitzmann2020a}.  NPs have been successfully applied for image completion \cite{garnelo2018, sitzmann2020a}, but to the best of our knowledge, this is the first work that successfully achieves conditional video prediction based on neural processes. In addition to VFP and VFI, the flexibility of NPs also enables our model to achieve video past frame extrapolation (VPE) and video random missing frames completion (VRC). By formulating conditional video prediction as a neural process, we build a supervised mapping from any target spatio-temporal coordinate of frames to target pixel values, given observed context coordinates and pixel values. The spatio-temporal coordinates are encoded by an implicit neural representation model to achieve continuous generation. More specifically, we firstly extract the features of each video frame by training a convolutional neural network (CNN) autoencoder. Then, a Transformer-based prediction model parameterizes an attentive neural process, which takes the target coordinates as inputs, conditions on context coordinates and context frame features, then outputs the target frame features that are finally fed into the CNN decoder to reconstruct the frame pixels. A Fourier Feature Network (FFN) learns the neural representations of coordinates, which serve as the positional information for the Transformer-based neural process model. Finally, a global latent variable is learned with variational methods to deal with prediction uncertainties. Our main contributions are:
\begin{itemize}
\item We propose the first neural process model for conditional video prediction (\textit{NPVP}), which tackles VFP, VFI, VPE and VRC with one model;

\item Our work is the first that successfully adapts INRs for temporal continuous VFP;

\item The proposed model is able to make temporal continuous video generation, i.e., generating video with an arbitrary high frame rate;

\item Our model outperforms the state-of-the-art (SOTA) models for VFP and VFI over multiple datasets.
\end{itemize}


\section{Background}
\noindent\textbf{Neural processes (NPs).} Given a set of labeled contexts $C = (X_C, Y_C) = \{(x_i, y_i)\}_{i\in \mathcal{I}(C)}$ and an unlabeled target set $T = X_T = \{x_i\}_{i\in \mathcal{I}(T)}$, Garnelo \etal \cite{garnelo2018} proposed (conditional) neural processes to model the predictive distribution $p(f(T)|C, T)$, where $\mathcal{I}(S)$ denotes the indices of data points in set $S$, function $f: X\rightarrow Y$ defines the mapping from domain $X$ to $Y$. Specifically, the contexts $C$ are firstly encoded and aggregated into a context embedding of fixed dimension, then $p(f(T)|C, T)$ is parameterized by a neural network with the inputs of context embedding and $T$. NPs are efficient because they preserve merits of both Gaussian processes and deep neural networks. An important property of NPs is that they are permutation invariant in $C$ and $T$ \cite{garnelo2018}. NPs can be extended to a latent variable version that accounts for the uncertainty of $f(T)$ based on VAE \cite{kingma2014}. In order to solve the underfitting problem of NPs, Kim \etal \cite{kim2019a} proposed to replace the context feature aggregation operation by an attention mechanism. NPs are required to be with scalability, flexibility and permutation invariance \cite{kim2019a}. They have been successfully applied for image completion \cite{garnelo2018, kim2019a, sitzmann2020a}. In this case, the pixel coordinates are considered as $x_i$ and pixel values are considered as $y_i$. Benefiting from the permutation invariance, NPs can predict missing pixel values condition on context pixels in arbitrary patterns. 

\noindent\textbf{Implicit neural representations (INRs).}
INRs \cite{tancik2020, sitzmann2020a} are techniques which solve the spectral bias problem of neural networks and thus achieve a continuous mapping between the input coordinates and target signal values, e.g., pixel values. There are mainly two different types of INRs. The first one is a Fourier Feature Network (FFN) \cite{tancik2020}, which uses a Fourier feature mapping for the input of a normal multiple layer perceptron (MLP) to enable the learning of high-frequency signal components effectively. The second one is a SInusoidal REpresentation Network (SIREN) \cite{sitzmann2020a}. SIREN depends on periodic activation functions, i.e., sinusoidal activations, to continuously represent the signals with fine details. Both FFN and SIREN are efficient, and some work  \cite{benbarka2022, yuce2022} have proven that they are equivalent to each other. INRs have been adopted for many computer vision tasks, including image generation \cite{skorokhodov2021a}, unconditional video generation \cite{skorokhodov2022} and video interpolation \cite{chen2022}.

\section{Related work}

Any video to video synthesis task can be considered as a conditional video prediction, including video translation between different domains \cite{wang2018}, video super-resolution \cite{haris2019, perez-pellitero2018}, VFP and VFI. We particularly focus on the work related to VFI and VFP. The classical supervised VFI models take optical flow-based \cite{jiang2018, niklaus2020} or kernel-based methods \cite{niklaus2017, niklaus2021} to learn the motion for the intermediate frames. The drawback is that those models require a high frame rate training dataset, which is relatively expensive to acquire. Some unsupervised VFI models have been proposed in recent years, for example, Reda \etal \cite{reda2019} developed a unsupervised VFI model based on cycle consistency. A more recent optical flow-based CNN model, VideoINR \cite{chen2022}, successfully utilizes the INRs for continuous VFI. 

VFP models can be categorized into many different types, such as deterministic models \cite{wu2020a, chen2020}, stochastic models \cite{babaeizadeh2018, denton2018}, pixel-direct generation models \cite{chen2020, franceschi2020} and transformation-based models \cite{chen2017, jin2018}. Almost all the VFP models are autoregressive models based on ConvLSTMs or Transformers \cite{yan2021a, wu2021b}. Recently, a few promising non-autoregressive VFP models were proposed \cite{liu2020e, ye2022, voleti2022}. VPTR \cite{ye2022} is a transformer-based non-autoregressive VFP model, but it only predicts future frames with a fixed frame rate. By combining ConvLSTMs with a neural ordinary differential equation (ODE) solver, Vid-ODE \cite{park2021b} is the first method that unifies the VFP and VFI into a single model, and it is able to generate temporally continuous video.  Another work, masked conditional video diffusion (MCVD) \cite{voleti2022} extends the 3D CNN-based diffusion models for video generation, but it is not an NP model and it is not able to do continuous video prediction. Benefiting from the flexibility of NPs, our model is able to perform video random missing frames completion (VRC) contrarily to MCVD. Furthermore, our model achieves stochastic prediction based on VAE instead of a diffusion model.

Our model is different from previous work in two aspects. Firstly, none of them are built to be a neural process. We believe that a NP is a better choice because it is permutation invariant in constrast to ConvLSTMs or 3D-CNN that are not. Therefore, they do not have the flexiblilty of our model to achieve multiple conditional video prediction tasks with a single model. Secondly, implicit neural representation (together with NPs) enables our model to predict frames at any given temporal coordinate, even though they are not seen during the training. Most of the previous models can only predict video frames with a fixed frame rate, which is defined by the training dataset. Vid-ODE \cite{park2021b} circumvents this limitation by introducing ODE. However, our NP-based model outperforms Vid-ODE for both VFP and VFI, as it will be shown in experiments. Another exception is VideoINR \cite{chen2022}, but VideoINR can only perform VFI and it does not satisfy the properties of NPs.

\section{Proposed method}
\begin{figure}
\centering
\includegraphics[clip, trim=7.2cm 12.3cm 6.8cm 10.9cm, width=0.88785\linewidth]{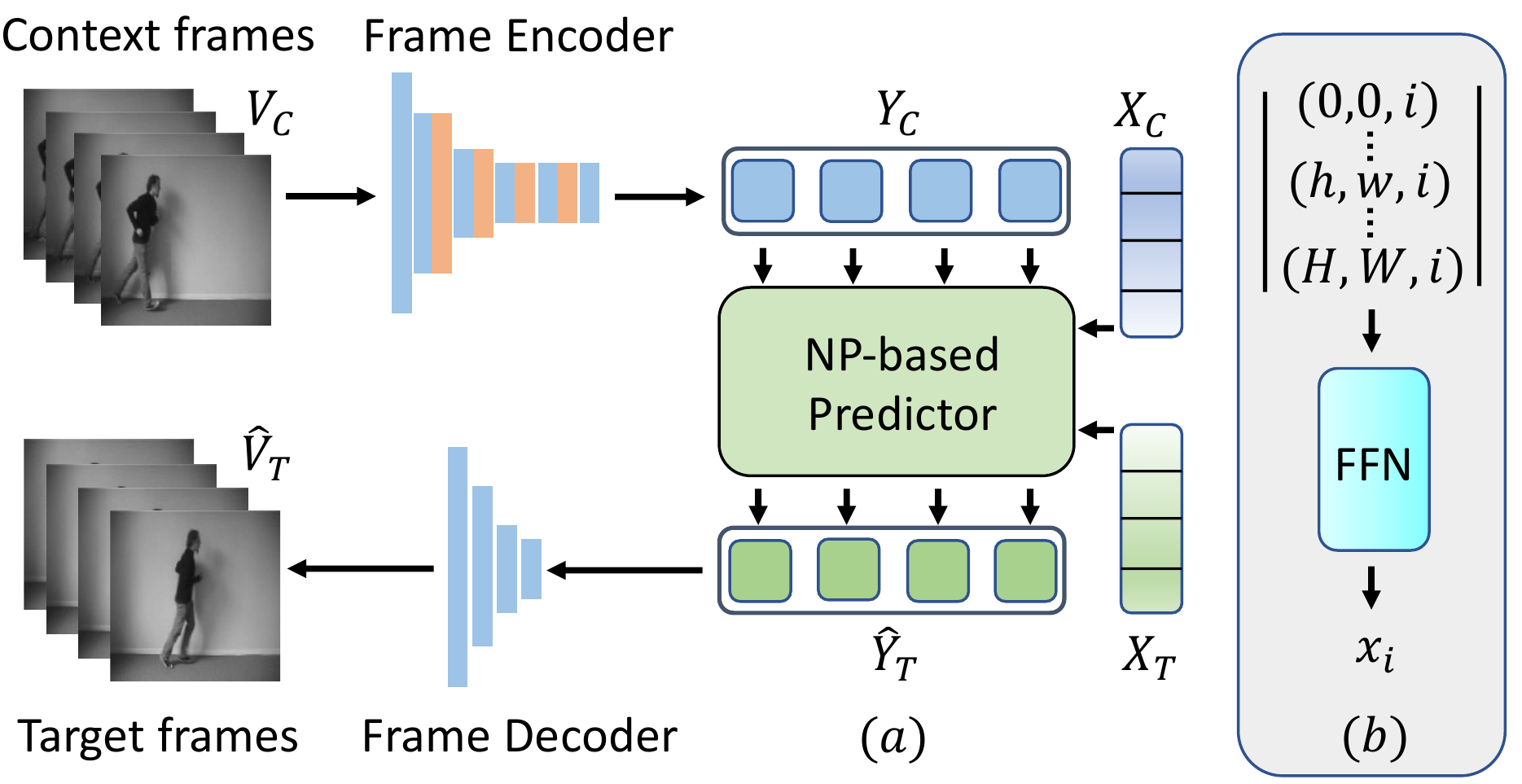}
\caption{(a) \textit{NPVP} framework. $Y_C$: context features; $\hat{Y}_T$: predicted target features; $X_C$/$X_T$: context/target spatio-temporal coordinate encodings.  (b) Implicit Neural Represenations (INRs).}
\label{fig:FrameworkINRs}
\end{figure}

Figure \ref{fig:FrameworkINRs} (a) depicts the proposed \textit{NPVP} framework. 
Given some context frames $V_C \in \mathbb{R}^{L_C\times I_h\times I_w \times I_c}$, \textit{NPVP} is trained to generate some target frames $V_T \in \mathbb{R}^{L_T\times I_h\times I_w \times I_c}$, where $L_C$ and $L_T$ denote the number of context frames and the number of target frames, respectively. $I_h, I_w, I_c$ are the image height, width, and the number of color channels. Firstly, the visual feature $Y_C \in \mathbb{R}^{L_C\times H\times W\times D}$ of context frames are extracted by a frame encoder, then a predictor predicts the target visual feature $\hat{Y}_T \in \mathbb{R}^{L_T\times H\times W\times D}$ given $Y_C, X_C$ and $X_T$. $X_C$ and $X_T$ denotes the context and target spatio-temporal coordinate encodings learned by a Fourier Feature Network (Figure \ref{fig:FrameworkINRs} (b)). $X_C\in \mathbb{R}^{L_C\times H\times W\times D}$ and $X_T \in \mathbb{R}^{L_T\times H\times W\times D}$, $H, W, D$ denote the visual feature height, width and channels. Finally, the target frames are reconstructed by a frame decoder given $\hat{Y}_T$.

Benefiting from the flexibility of NPs, given a video clip, we can randomly select any number of frames at any time step $i$ to be the context frames $V_C$, and the remaining ones are the frames $V_T$ to be predicted. In this way, our model can be trained as a general model for multiple conditional video prediction tasks. For example, if we make all the context frames have a smaller time coordinates than all the target frames, the model is trained for VFP specifically, but if a model is trained with random context, it is able to solve multiple conditional prediction tasks at the same time.

We divided our model into two parts, a frame autoencoder and a NP-based predictor that operates over the feature space. The reason for our choice is that directly learning a NP-based model over the pixel space is expensive. Operating over the feature space allows us to train the model in two stages. We firstly train the frame encoder and frame decoder by ignoring the NP-based predictor, and then fix the parameters of the frame autoencoder to learn the predictor. The detail architectures of the autoencoder, Fourier feature network (Figure \ref{fig:FrameworkINRs} (b)), and the NP-based Predictor (Figure \ref{fig:Predictor}) are described in the following.

\subsection{Autoencoder}
We use a custom autoencoder that is adapted from Pix2Pix \cite{Isola2017}. Specifically, we integrate non-local 2d attention layers (orange layers of the frame encoder in Figure \ref{fig:FrameworkINRs} (a)) from SAGAN \cite{zhang2019b} into the CNN encoder to improve its performance. There is no modification for the frame decoder of Pix2Pix. The autoencoder is trained with a simple $L_1$ loss between input frame $I$ and reconstructed frame $\hat{I}$, i.e., $\mathcal{L}_{1}(I, \hat{I}) = \lvert I - \hat{I}\rvert$. Recall that the predictor is ignored during the learning of the autoencoder, and the autoencoder is fixed during the learning of the predictor.

\subsection{Fourier feature network for INRs}
We selected a FFN \cite{tancik2020} instead of SIREN \cite{sitzmann2020a} because a FFN is easier to train.  For a visual feature $y_i \in \mathbb{R}^{H\times W\times D}$ of one frame, where $i$ is the temporal coordinate, the FFN takes the coordinate $(h, w, i)$ of each feature vector at different spatio-temporal location as input, and outputs a $D$-dimensional coordinate encoding for $(h, w, i)$. The INRs is shown Figure \ref{fig:FrameworkINRs} (b). $x_i \in \mathbb{R}^{H\times W\times D}$ denotes all the spatial-temporal coordinate encodings of a frame feature $y_i$. Then, $X_C$ and $X_T$ contains all the $x_i$ of the context and target coordinates respectively. Specifically, for an input 3D coordinate vector $(h, w, i)$, the FFN firstly projects it to a higher dimensional space by a Gaussian random noise matrix, then the projections are fed into a MLP with ReLU activation functions to get the output coordinate encoding. The spatio-temporal coordinates are normalized to the range $[0, 1]$. The FFN is jointly learned with the NP-based predictor. 

The implicit neural representations $X_C$ and $X_T$ generated by the FFN encode the spatio-temporal location information of context features $Y_C$ and target features $Y_T$. They are critical for the learning of our Transformer-based predictor (section \ref{ssec: predictors}) because a Transformer is permutation invariant. After training, INRs are able to generalize to unseen input coordinates, which means that we can get the coordinate encoding $x_i$ at any real-number temporal coordinate $i$. Because the model predicts different $y_i$ given contexts and different target $x_i$ (section \ref{ssec: predictors}), we can achieve a continuous generation. For the VFP task, if we need to predict target frames beyond the maximum temporal coordinates used during training, we can perform a "block-wise" autoregressive prediction. Specifically, take the predicted future (target) frames as the past (context) frames for the next block of future (target) frames.

\subsection{NP-based Predictor}
\label{ssec: predictors}

Our NP-based predictor is designed as an attentive neural process \cite{kim2019a} based on a video representation learning Transformer, VidHRFormer \cite{ye2022}. This is motivated by the fact that an attentive neural process preserves the permutation invariance and solves the underfitting problem of a vanilla NP, and VidHRFormer satisfies all the requirements of attentive neural processes.

\noindent\textbf{Loss function.} Given contexts $(X_C, Y_C)$ and $X_T$, a NP learns to maximize conditional log-likelihood $\log p(Y_T|X_C,Y_C,X_T)$. In other words, a NP makes probabilistic predictions for $Y_T$, which is normally assumed to follow a factorized Gaussian distribution $p$ \cite{garnelo2018, kim2019a}. However, we argue that a simpler point prediction is better for video prediction. Firstly, $Y_T$ has a much higher dimensionality than the simple regression datasets or images in \cite{garnelo2018, kim2019a}, therefore it is expensive to predict the covariance even if it is diagonal (factorized). Preliminary experiments show that the predicted diagonal covariance only captures the variation of high-frequency noise instead of the desired temporal uncertainty of motion. Furthermore, even if the predicted diagonal covariance successfully estimates the motion uncertainty at each time step, we cannot use it to sample coherent video sequences because each time step is independent of each other \cite{garnelo2018}. Considering that learning full covariance is infeasible, one solution is to enforce causality among different time steps by making autoregressive predictions, which however suffers from accumulated errors and low inference speed.

Therefore, we propose a better solution which is to model $p$ as a Laplacian distribution with a constant scale parameter. This corresponds to an efficient point prediction. In order to achieve coherent sequence sampling, we introduce a global latent variable $z_e$ into the vanilla NP \cite{garnelo2018}, where $z_e$ explains the uncertainty of the whole sequence $Y_T$ and thus is named "event variable". Thus, we can describe the generative process of $Y_T$ as:
\begin{multline}
    p(Y_T|X_C, Y_C, X_T) = \\
    \int p(Y_T|X_T, X_C, Y_C, z_e)q(z_e|X_C, Y_C)dz_e,
\label{eq:generative_process}
\end{multline}
where $q(z_e|X_C, Y_C)$ defines a conditional prior distribution for $z_e$. By adapting a VAE \cite{kingma2014}, we can approximate Eq. \ref{eq:generative_process} by maximizing the evidence lower bound (ELBO):
\begin{multline}
    ELBO = \mathbb{E}_{q_\phi(z_e|X_T, Y_T)}[\log p(Y_T|X_T, X_C, Y_C, z_e)] \\
    - \beta D_{KL}(q_{\phi}(z_e|X_T, Y_T)||q_{\psi}(z_e|X_C, Y_C)),
\label{eq:ELBO}
\end{multline}
where $\beta$ is a hyperparameter. $\phi$ and $\psi$ denote the parameters of two factorized Gaussian distribution $\mathcal{N}(\mu_\phi(X_T, Y_T), \sigma_\phi(X_T, Y_T))$ and $\mathcal{N}(\mu_\psi(X_C, Y_C), \sigma_\psi(X_C, Y_C))$ respectively. Specifically, the first term in the RHS of Eq. \ref{eq:ELBO} forces the predictor to reconstruct $Y_T$. The KL divergence is a regularization term that prevents the $z_e$ sampled from targets to deviate too far from the $z_e$ sampled from the context, with the assumption that both context frames and target frames are generated from the same latent event space. 

As $p$ follows a Laplacian distribution with a constant scale parameter, maximizing the log-likelihood (first term in the RHS of Eq. \ref{eq:ELBO}) is equivalent to minimizing an $L_1$ loss, that is $\lvert Y_T - \hat{Y}_T\rvert$. We can derive the loss function as 

\begin{equation}
    \mathcal{L} = \lvert Y_T - \hat{Y}_T\rvert + \beta D_{KL}(q_{\phi}(z_e|X_T, Y_T)||q_{\psi}(z_e|X_C, Y_C)).
\label{eq:loss_simple}
\end{equation}

However, in practice, we find that learning with Eq. \ref{eq:loss_simple} cannot generate predictions with good visual quality, because the $L1$ loss in the RHS of Eq. \ref{eq:loss_simple} does not consider the curvature of the latent feature manifold learned by the frame autoencoder \cite{shao2018}. Therefore, we feed $\hat{Y}_T$ to the fixed frame decoder to reconstruct target frames $\hat{V}_T$ and minimize another pixel reconstruction $L_1$ loss, i.e., $\lvert V_T - \hat{V}_T\rvert$, at the same time. In this way, the supervisory signal from the pixel $L_1$ loss minimizes the geodesic distance between $Y_T$ and $\hat{Y}_T$ \cite{bhagat2020b}. Then, the final loss function is
\begin{multline}
    \mathcal{L} = \gamma \lvert V_T - \hat{V}_T\rvert + \lvert Y_T - \hat{Y}_T\rvert \\ + \beta D_{KL}(q_{\phi}(z_e|X_T, Y_T)||q_{\psi}(z_e|X_C, Y_C)),
\label{eq:loss_final}
\end{multline}
where $\gamma$ is a hyperparameter.

\begin{figure}
\centering
\includegraphics[clip, trim=8cm 9.5cm 3.6cm 9.6cm, width=\linewidth]{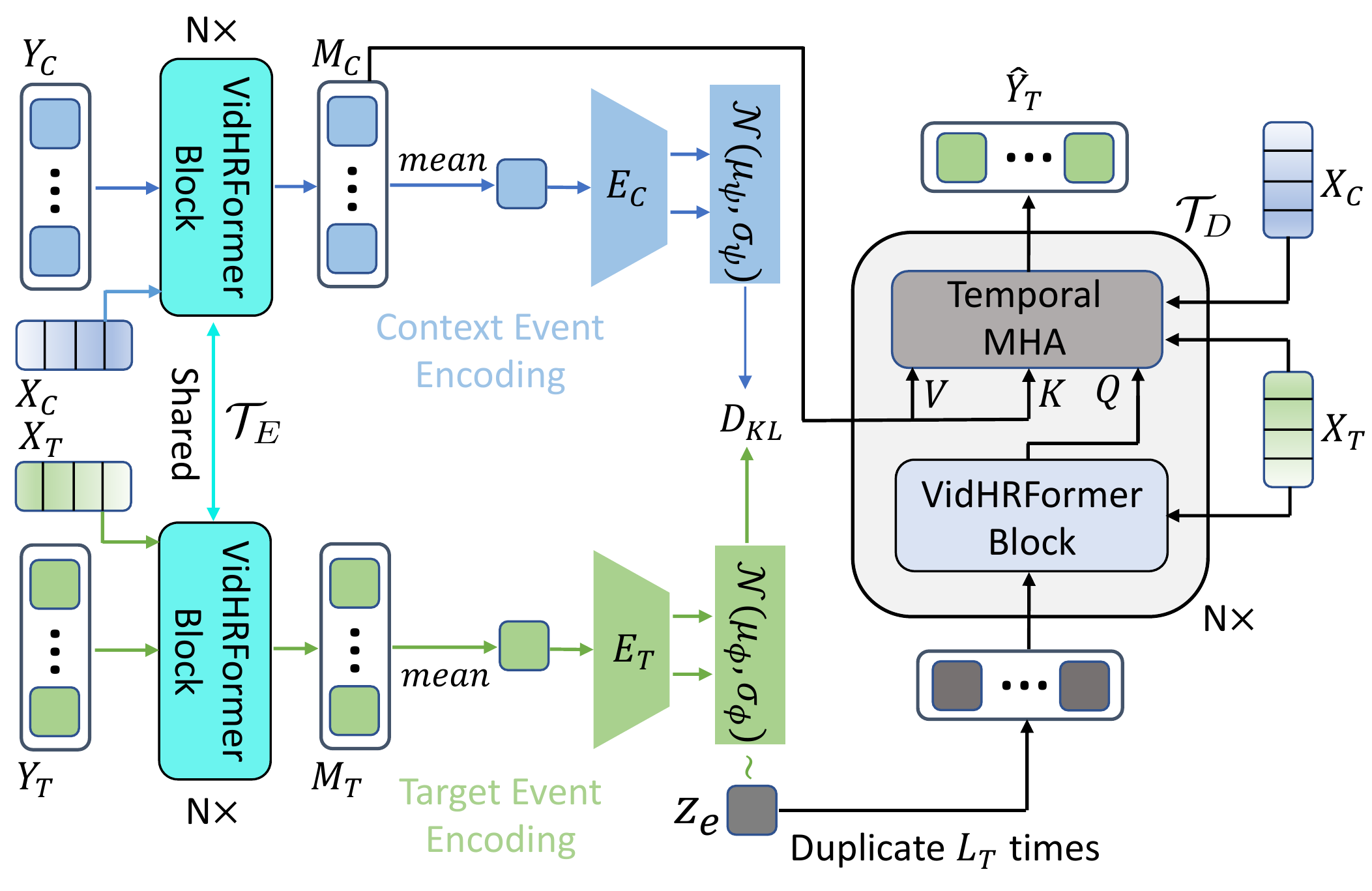}
\caption{Architecture of the NP-based predictor. Target event encoding is only used during training. $z_e$ is sampled from $\mathcal{N}(\mu_\psi, \sigma_\psi)$ during test.}
\label{fig:Predictor}
\end{figure}

\noindent\textbf{Network architecture.} The architecture of the proposed NP-based predictor is shown in Figure \ref{fig:Predictor}. It is composed of a context event encoding module, a target event encoding module and a Transformer Decoder $\mathcal{T}_D$. The context event encoding shares the same architecture as the target event encoding, they are responsible for learning $q_\psi$ and $q_\phi$ respectively. $\mathcal{T}_D$ is responsible for predicting $\hat{Y}_T$. 

In detail, the architecture of the context event encoding path can be formalized by these operations,
\begin{align}
M_C &= \mathcal{T}_E(X_C, Y_C) \\
\mu_\psi, \sigma_\psi &= E_C(mean(M_C)),
\label{eq:context_event_encoding}
\end{align}

\noindent where $\mathcal{T}_E: X_C \times Y_C \rightarrow M_C \in \mathbb{R}^{L_C\times H\times W\times D}$ denotes a Transformer encoder, which is composed by multiple VidHRFormer blocks \cite{ye2022}. The spatio-temporal separated attention mechanism of VidHRFormer block ensures the permutation invariance along the temporal dimension. $X_C$ is fused into $\mathcal{T}_E$ as the positional encodings. In order to generate $(\mu_\psi, \sigma_\psi)$, $M_C$ is firstly averaged along the temporal dimension, then passed through an event encoder $E_C: \mathbb{R}^{H\times W\times D} \rightarrow \mathbb{R}^{H\times W\times D}$, where $E_C$ is a small CNN with two output heads for $\mu_\psi$ and $\sigma_\psi$, respectively. We use the $mean$ operation because it is an efficient aggregation method and it is permutation invariant. Please see the supplementary material for the detailed architectures of VidHRFormer block \cite{ye2022} and $E_C$.

The target event encoding path has the same architecture as the context event encoding path, which is formalized as,
\begin{align}
M_T &= \mathcal{T}_E(X_T, Y_T) \\
\mu_\phi, \sigma_\phi &= E_T(mean(M_T)),
\label{eq:target_event_encoding}
\end{align}
where $M_T \in \mathbb{R}^{L_T\times H\times W\times D}$. Note that the target event encoding shares the same $\mathcal{T}_E$ with the context event encoding, but it has a different target event encoder $E_T: \mathbb{R}^{H\times W\times D} \rightarrow \mathbb{R}^{H\times W\times D}$.

We hypothesize that all target visual features are generated by event variable $z_e \in \mathbb{R}^{H\times W\times D}$, and $z_e \sim \mathcal{N}(\mu_\phi, \sigma_\phi)$ during training. During test, the ground truth $Y_T$ is not accessible, then $z_e$ is sampled from the learned context prior event space $\mathcal{N}(\mu_\psi, \sigma_\psi)$, which is generated by the context event encoding path.

Finally, $\hat{Y}_T$ is generated by conditioning on $(X_T, X_C, M_C, z_e)$ through another Transformer $\mathcal{T}_D$,
\begin{equation}
    \hat{Y}_T = \mathcal{T}_D(X_T, X_C, M_C, z_e).
\label{eq: Transformer_decoder}
\end{equation}

The architecture of a $\mathcal{T}_D$ block is the same as the Transformer decoder block of used in VPTR \cite{ye2022} (see supplementary material for the detailed architecture). $M_C$ and $X_C$ are fed into $\mathcal{T}_D$ as the key/value and positional encodings of the encoder-decoder temporal multi-head attention layer respectively. They provide context information for $\hat{Y}_T$. $X_T$ is also injected into $\mathcal{T}_D$ as positional encodings. Note that event variable $z_e$ is duplicated $L_T$ times and fed into $\mathcal{T}_D$ as the initial query for each target frame feature. In this way, we can generate $\hat{Y}_T$ with arbitrary frame rate, i.e., continuous prediction, as long as we input the desired $X_T$, which is produced by the trained FFN for free. 

\begin{table*}[t]
\setlength{\tabcolsep}{1pt}
\centering
\resizebox{\textwidth}{!}{\begin{tabular}{cccccc} \hline
\multicolumn{3}{c}{\multirow{2}{*}{Models}} & \multicolumn{3}{c}{KTH, \textit{10 $\rightarrow$ 20}}\\
& & & PSNR$\uparrow$ & SSIM$\uparrow$ & LPIPS $\downarrow$ \\ 
\hline
\multicolumn{3}{c}{PredRNN++ \cite{wang2018d}} & 28.47 & 0.865 & 228.9\\
\multicolumn{3}{c}{STMFANet \cite{jin2020}}& \textbf{29.85} & 0.893 & 118.1 \\
\multicolumn{3}{c}{E3D-LSTM \cite{wang2018a}} & \textcolor{blue}{\textit{29.31}} & 0.879 & - \\
\multicolumn{3}{c}{Conv-TT-LSTM \cite{su2020a}} & 28.36 & \textcolor{blue}{\textit{0.907}} & 133.4 \\
\multicolumn{3}{c}{Vid-ODE \cite{park2021b}}& 28.19 & 0.878 & \textcolor{blue}{\textit{80.0}}\\
\multicolumn{3}{c}{VPTR-NAR \cite{ye2022}} & 26.96 & 0.879 & 86.1 \\ \hline
\multicolumn{3}{c}{\textit{NPVP} (ours)}& 27.66 & \textbf{0.909} & \textbf{66.0} \\
\hline
\end{tabular}
\quad
\begin{tabular}{ccccc} \hline
\multicolumn{3}{c}{\multirow{2}{*}{Models}} & \multicolumn{2}{c}{KITTI, \textit{4 $\rightarrow$ 5}} \\
& & & SSIM$\uparrow$ & LPIPS$\downarrow$ \\ \hline
\multicolumn{3}{c}{PredNet \cite{lotter2017}} & 47.56 & 629.5\\
\multicolumn{3}{c}{MCNet\cite{villegas2017a}}& 55.48 & 373.9\\
\multicolumn{3}{c}{Voxel Flow \cite{liu2017} } & 42.62 & 415.9\\
\multicolumn{3}{c}{FVS \cite{wu2020a}} & 60.77 & \textcolor{blue}{304.9} \\
\multicolumn{3}{c}{SADM \cite{bei2021}} & \textcolor{blue}{64.72} & 311.6 \\ \hline
\multicolumn{3}{c}{\textit{NPVP} (ours)} & \textbf{66.12} & \textbf{279.0} \\ \hline
\end{tabular}
\quad
\begin{tabular}{cccccc} \hline
\multicolumn{3}{c}{\multirow{2}{*}{Models}} & \multicolumn{3}{c}{Cityscapes, \textit{2 $\rightarrow$ 28}} \\
& & & FVD$\downarrow$ & SSIM$\uparrow$ & LPIPS$\downarrow$ \\ 
\hline

\multicolumn{3}{c}{SVG-LP \cite{denton2018}} & 1300.26 & 0.574 & 549.0 \\ 
\multicolumn{3}{c}{VRNN 1L\cite{castrejon2019}}& 682.08 & 0.609 & 304.0 \\
\multicolumn{3}{c}{Hier-VRNN \cite{castrejon2019}}& 567.51 & 0.628 & 264.0 \\
\multicolumn{3}{c}{GHVAEs \cite{wu2021a}}& \textcolor{blue}{\textit{418.00}}& \textcolor{blue}{\textit{0.740}} & 194.0 \\
\multicolumn{3}{c}{MCVD-concat \cite{voleti2022}} & \textbf{141.31} & 0.690 & \textbf{112.0} \\
\hline
\multicolumn{3}{c}{\textit{NPVP} (ours)} & 768.04 & \textbf{0.744} & \textcolor{blue}{\textit{183.2}}\\
\hline
\end{tabular}}

\caption{VFP results on KTH, KITTI and Cityscapes. \textbf{Boldface}: best results. \textit{\textcolor{blue}{Blue}}: second best results.}
\label{tab:VFP_results}
\end{table*}

\section{Experiments}
We evaluated the proposed predictor on multiple realistic video datasets, KTH \cite{schuldt2004}, BAIR \cite{ebert2017}, KITTI \cite{geiger2013}, Cityscapes \cite{cordts2016}, and a synthetic video dataset, Stochastic Moving MNIST (SM-MNIST) \cite{denton2018}. KITTI and Cityscapes are resized to $128\times 128$, all others are resized to the resolution of $64\times 64$. Following the experimental configuration of previous work, we present the quantitative results of Peak Signal-to-Noise Ratio (PSNR), Fréchet Video Distance (FVD)\cite{unterthiner2019b}, Learned Perceptual Image Patch Similarity (LPIPS)\cite{zhang2018} and Structural Similarity Index Measure (SSIM). The LPIPS is reported in $10^{-3}$ scale. Same as previous stochastic methods, 100 different predictions are sampled for each test example, then the best SSIM, LPIPS, PSNR, and the average FVD of the generated samples are reported.

For a fair comparison with previous task-specific models, we first train different models for VFP and VFI respectively, i.e., we train task-specific models following the same training procedures that they used. Then, we present results with a unified model that is not task-specific, but since the training is different, results are not fully comparable with task-specific models. The supplementary material includes more qualitative examples and implementation details.

\subsection{Task-specific models}

\begin{figure}
\centering
\includegraphics[clip, trim=8.5cm 12cm 9cm 8cm, width=\linewidth]{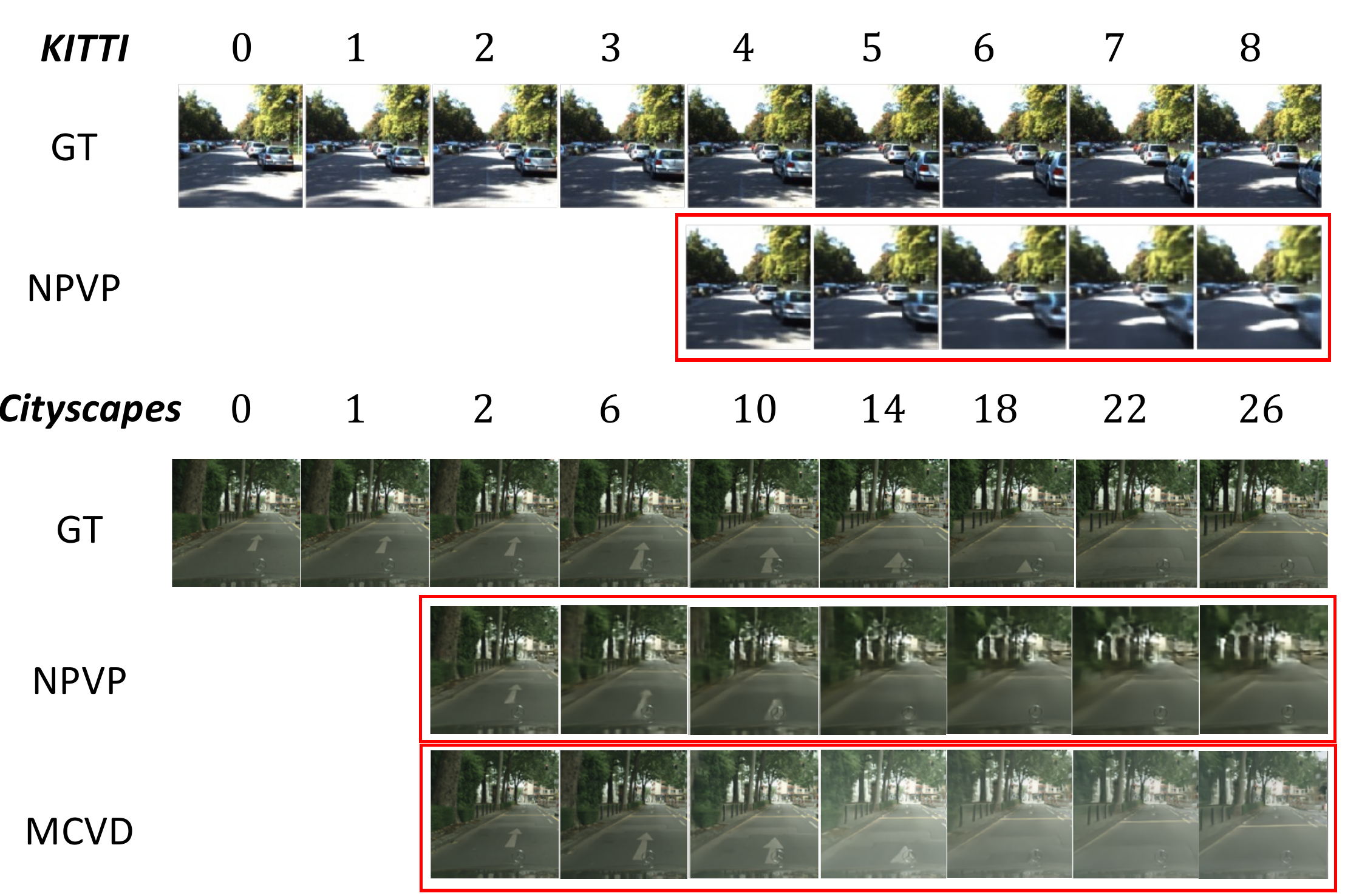}
\caption{VFP examples on KITTI and Cityscapes. Frames inside the red boxes are future frames predicted
by the model. }
\label{fig:VFP_examples}
\end{figure}

\noindent\textbf{VFP.} The VFP experimental results are summarized in Table \ref{tab:VFP_results}. For the KTH dataset, our \textit{NPVP} model is trained to predict 10 future frames given 10 past frames. During test, the performance is evaluated on predicting 20 future frames conditioned on 10 past frames, which is achieved by a block-wise autoregressive inference. Our \textit{NPVP} achieves the best SSIM and outperforms previous methods by a large margin in terms of LPIPS. For KITTI, \textit{NPVP} is trained to predict 5 future frames given 4 past frames. Compared with previous methods, our \textit{NPVP} reaches the best performance in terms of both SSIM and LPIPS. Qualitative results (Figure \ref{fig:VFP_examples}) show that \textit{NPVP} predicts future frames with good visual quality despite the large motion of KITTI dataset, which has a low frame rate of 10 fps. The results on KITTI dataset demonstrate that \textit{NPVP} is capable of challenging real-world traffic video prediction.

For the Cityscapes dataset,  \textit{NPVP} is trained to predict 10 future frames given 2 past frames, but 28 future frames are predicted by block-wise autoregressive inference during test. \textit{NPVP} achieves the best SSIM and the second-best LPIPS. We suspect that the gap between \textit{NPVP} and MCVD-concat in terms of FVD is due to the fact that a vanilla VAE is not expressive enough \cite{castrejon2019}. All methods for Cityscapes in Table \ref{tab:VFP_results} are VAE-based, except for the MCVD-concat, which uses a denoising diffusion model (DDM) \cite{ho2020, song2021}. It has a brightness-changing problem \cite{voleti2022}, but outperforms all VAE-based methods for the FVD score. Castrejón \etal \cite{castrejon2019} has shown that increasing the levels of VAE latent variables is beneficial for the expressiveness of VFP models. Meanwhile, DDM can be considered as a special form of hierarchical VAEs with a large levels of latent variable, and all the latent variables have the same dimensionality as the data, i.e., a video clip. Therefore, MCVD-concat achieves the best FVD probably because of the high expressiveness of its latent variables. Our proposed method may suffer from weaker temporal coherence caused by its non-autoregressive prediction. All VAE-based models in Table \ref{tab:VFP_results} are auto-regressive models, but the loss function of \textit{NPVP} assumes that frames at different time steps are independent of each other to preserve the unified model flexibility, even though the temporal attention exchange information between different frames. Nevertheless, \textit{NPVP} achieves a comparable or better performance than the SOTAs. Visual examples of VFP on Cityscapes (Figure \ref{fig:VFP_examples}) shows that our method predicts better than MCVD \cite{voleti2022} that suffers from brightness change problems.


\begin{table*}[t]
\setlength{\tabcolsep}{1pt}
\centering
\begin{tabular}{ccccccc|cccc|cccc} \hline
\multicolumn{3}{c}{\multirow{2}{*}{Models}} & \multicolumn{4}{c}{KTH} & \multicolumn{4}{c}{SM-MNIST} & \multicolumn{4}{c}{BAIR} \\
& & & ($p$+$f$$\rightarrow$$k$) & PSNR$\uparrow$ & SSIM$\uparrow$ & LPIPS$\downarrow$ &  ($p$+$f$$\rightarrow$$k$) & PSNR$\uparrow$ & SSIM$\uparrow$ & LPIPS$\downarrow$ & ($p$+$f$$\rightarrow$$k$) & PSNR$\uparrow$ & SSIM$\uparrow$ & LPIPS$\downarrow$\\ 
\hline
\multicolumn{3}{c}{SVG-LP \cite{denton2018}}& (\textit{18$\rightarrow$7}) & 28.13 & 0.883 & - & (\textit{18$\rightarrow$7}) & 13.54 & 0.741 & - & (\textit{18$\rightarrow$7}) & 18.65 & 0.846 & - \\
\multicolumn{3}{c}{SDVI-\textit{full} \cite{xu2020a}}& (\textit{18$\rightarrow$7}) & 29.19 & 0.901 & - &  (\textit{18$\rightarrow$7}) & 16.03 & 0.842 & - & (\textit{18$\rightarrow$7}) & 21.43 & 0.880 & - \\
\multicolumn{3}{c}{Vid-ODE \cite{park2021b}}& - & 31.77 & 0.911 & 48.0 & - & - & - & - & - & - & - & - \\
\multicolumn{3}{c}{MCVD \cite{voleti2022}}& (\textit{15$\rightarrow$10}) & 34.67 & 0.943 & - &  (\textit{10$\rightarrow$10}) & 20.94 & 0.854 & - & (\textit{4$\rightarrow$5}) & \textcolor{blue}{\textit{25.16}} & \textcolor{blue}{\textit{0.932}} & - \\ 
\multicolumn{3}{c}{MCVD \cite{voleti2022}}& (\textit{10$\rightarrow$5}) & \textcolor{blue}{\textit{35.61}} & 0.963 & - & (\textit{10$\rightarrow$5}) & 27.69 & 0.940 & - & - & - & - & - \\ \hline

\multicolumn{3}{c}{\multirow{2}{*}{\textit{NPVP (ours)}}}& (\textit{15$\rightarrow$10})$\dagger$ & 33.60 & \textcolor{blue}{\textit{0.969}} & \textcolor{blue}{\textit{22.3}} & (10$\rightarrow$10) & \textcolor{blue}{\textit{28.11}} & \textcolor{blue}{\textit{0.958}} & \textcolor{blue}{\textit{17.3}} & (\textit{18$\rightarrow$7}) & 22.97 & 0.909 & \textcolor{blue}{\textit{21.8}} \\
& & & (10$\rightarrow$5) & \textbf{37.17} & \textbf{0.984} & \textbf{10.5} & (10$\rightarrow$5) & \textbf{34.34} & \textbf{0.992} & \textbf{4.1} & (\textit{4$\rightarrow$5}) & \textbf{25.28} & \textbf{0.933} &  \textbf{14.7} \\
\hline

\end{tabular}
\caption{VFI results. $p/f$: number of past$/$future frames; $k$: number of intermediate frames to interpolate. \textbf{Smaller $p$+$f$ and larger $k$ means a harder VFI task}. $\dagger$: $p=8$, $f=7$; for our other models, $p$ equals to $f$. \textbf{Boldface}: best results. \textit{\textcolor{blue}{Blue}}: second best results.}
\label{tab:VFI_results}
\end{table*}

\noindent\textbf{VFI.} We follow the experimental protocol in \cite{voleti2022}. For the VFI task, given $p$ past frames and $f$ future frames as the context, the model is trained to generate $k$ intermediate frames, i.e., target frames. For VFI, the context spans across the past and future, which limits the event stochasticity.

The VFI results are summarized in Table \ref{tab:VFI_results}. For the KTH dataset, our task-specific \textit{NPVP} (\textit{15$\rightarrow$10}) model outperforms the SOTA MCVD (\textit{15$\rightarrow$10}) in terms of SSIM. It also outperforms Vid-ODE in terms of all metrics by a large margin. Note that a smaller $p+f$ and a larger $k$ means a harder VFI task. Therefore, we can draw the conclusion that our model is better than SVG-LP and SDVI-\textit{full}, because (\textit{15$\rightarrow$10}) is harder than (\textit{18$\rightarrow$7}). Our \textit{NPVP} (\textit{10$\rightarrow$5}) also outperforms MCVD (\textit{10$\rightarrow$5}) model in terms of both SSIM and PSNR. The \textit{NPVP} (\textit{10$\rightarrow$5}) outperforms \textit{NPVP} (\textit{15$\rightarrow$10}). We believe that it is because 10 context frames are enough to provide good context information for the KTH dataset, and interpolating 5 frames is much easier than interpolating 10 frames.

For the SM-MNIST dataset, our \textit{NPVP} (\textit{10$\rightarrow$10}) outperforms all the previous methods in terms of both PSNR and SSIM by a large margin, even for the easier (\textit{18$\rightarrow$7}) and (\textit{10$\rightarrow$5}) tasks. It means that given the same number of context frames, our model is able to interpolate more intermediate frames with better quality. The \textit{NPVP} (\textit{10$\rightarrow$5}) outperforms \textit{NPVP} (\textit{10$\rightarrow$10}) as expected. In short, the results demonstrate that \textit{NPVP} achieves a new SOTA for VFI on the SM-MNIST dataset. The randomness of SM-MNIST only occurs when the characters bounce off the boundaries. Most of the time, the character trajectories of intermediate frames can be determined based on the past and future frames, thus the model tends to ignore the event variable as little stochasticity exists. A similar phenomenon is observed for VFI on the KTH dataset because natural human motion in missing frames is mostly constrained by past and future movements.

For the BAIR dataset, our \textit{NPVP} (\textit{18$\rightarrow$7}) outperforms SVG-LP and SDVI-\textit{full} by a large margin. Compared with MCVD (\textit{4$\rightarrow$5}), \textit{NPVP} (\textit{4$\rightarrow$5}) achieves a slightly better performance in terms of both PSNR and SSIM. 

\subsection{A unified model for VFP, VFI, VPE and VRC}
\label{sec:random_context}
We trained a unified \textit{NPVP} on the KTH dataset to demonstrate that our NP-based conditional video prediction model is flexible enough to perform VFP, VFI, video past frame extrapolation (VPE), and video random missing frames completion (VRC) with one single learned model. More importantly, all the aforementioned tasks can be solved with an arbitrary high frame rate, i.e., a continuous prediction. In order to learn one model for all tasks, \textit{NPVP} is trained with random contexts, i.e., given a video clip with $L$ frames, we draw $L_C$ frames (the probability of each frame follows a uniform distribution) as contexts and the remaining $L_T = L-L_C$ frames are target frames, together with their corresponding coordinates. $L = 20$, and the value of $L_C$ varies in the range of $[4, 16]$. 

\begin{figure}
\centering
\includegraphics[clip, trim=5cm 16.6cm 10.1cm 2cm, width=\linewidth]{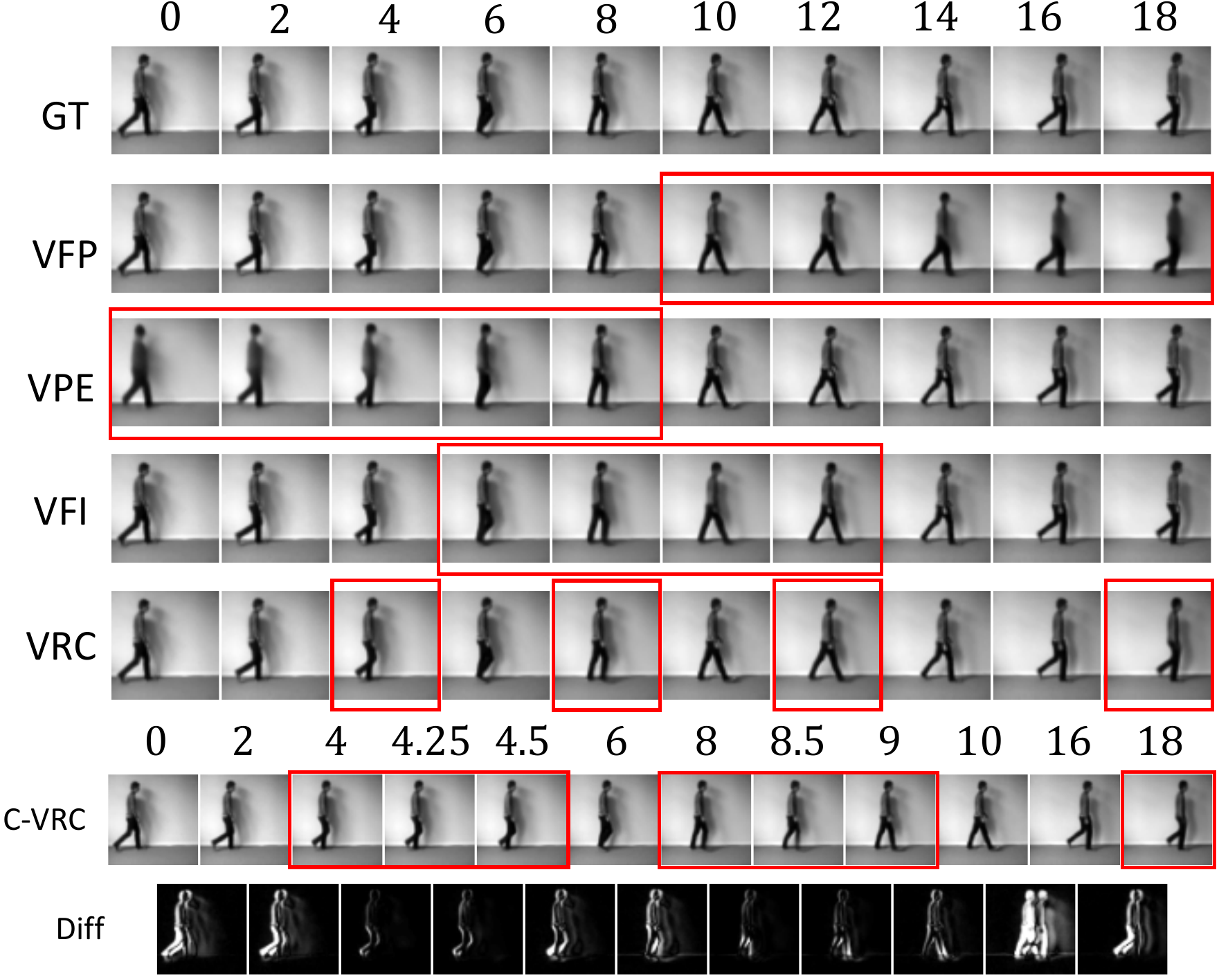}
\caption{One model for all tasks. Frames inside the red boxes are target frames generated by the model. C-VRC denotes continuous VRC. Diff are the difference images between neighboring frames of C-VRC to show that they are all different and that the temporal coordinates are taken into account. }
\label{fig:OneForAll}
\end{figure}

\begin{figure}
\centering
\includegraphics[width=\linewidth]{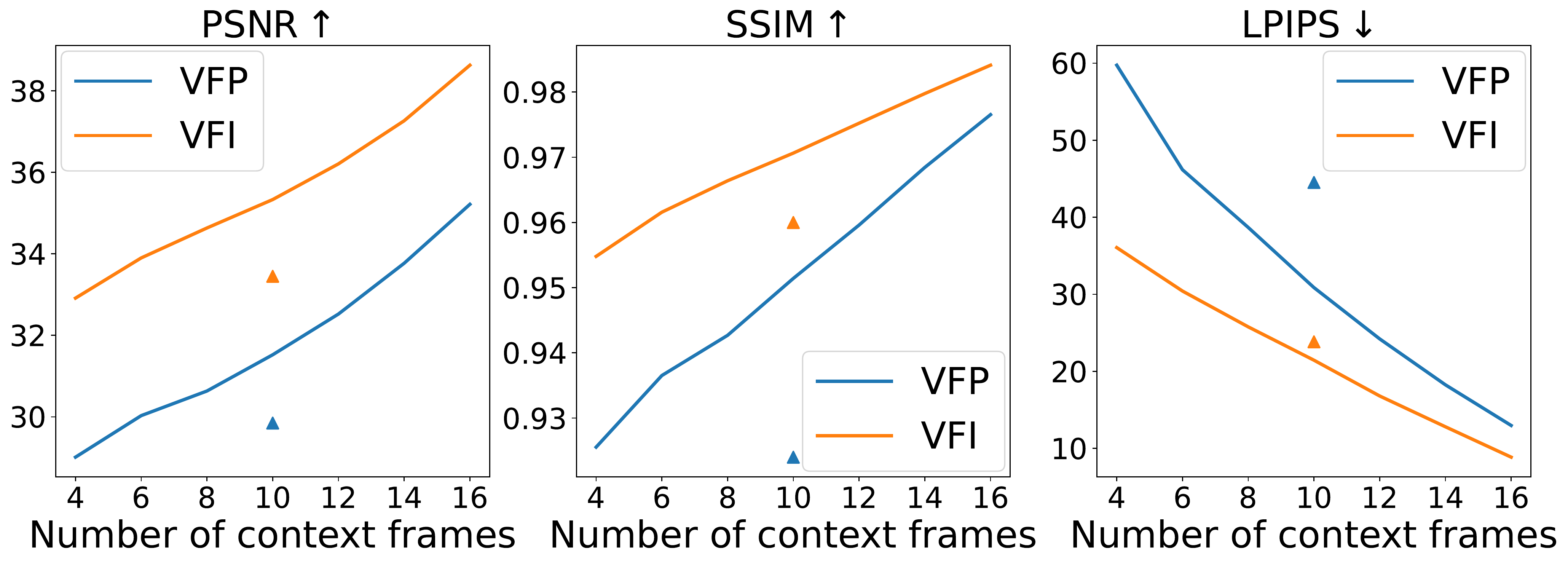}
\caption{Metric curves of VFP and VFI on KTH for an increasing number of context frames. \textcolor{orange}{$\blacktriangle$} and \textcolor{RoyalBlue}{$\blacktriangle$} denote results of task-specific \textit{NPVP} (\textit{10$\rightarrow$10}) for VFI and VFP respectively.}
\label{fig:varying_contexts}
\end{figure}

\setlength{\tabcolsep}{2pt}
\begin{table*}[t]
\centering
\begin{tabular}{c|ccccccc|ccc|ccc} \hline
& \multicolumn{6}{c}{Models} & \multicolumn{3}{c}{VFI, \textit{5+5 $\rightarrow$10}} & \multicolumn{3}{c}{VFP, \textit{10 $\rightarrow$ 10}} \\
& INR & $\mathcal{T}_D$-$6L$ & $\mathcal{T}_D$-$8L$ & $\mathcal{T}_E$ & $FL_1$ & $PL_1$ & $D_{KL}$ & PSNR$\uparrow$ & SSIM$\uparrow$ & LPIPS$\downarrow$ & PSNR$\uparrow$ & SSIM$\uparrow$ & LPIPS$\downarrow$ \\ 
\hline
1 & & \checkmark & & & \checkmark & & & 30.58 & 0.937 & 50.07 & 28.05 & 0.904 & 71.75 \\
2 & \checkmark & \checkmark & & & \checkmark & & & 31.34 & 0.953 & 37.03 & 28.83 & 0.920 & 71.04 \\
3 & \checkmark & \checkmark & & & \checkmark & \checkmark & & 33.17 & 0.958 & 31.11 & 29.60 & 0.920 & 62.96 \\
4 & \checkmark & & \checkmark & & \checkmark & \checkmark & & 33.20 & 0.959 & 29.64 & 29.85 & 0.922 & 57.49 \\
5 & \checkmark & & \checkmark & \checkmark & \checkmark & \checkmark & & \textcolor{blue}{\textit{33.77}} & \textcolor{blue}{\textit{0.962}} & \textcolor{blue}{\textit{26.83}} & \textcolor{blue}{\textit{30.11}} & \textcolor{blue}{\textit{0.927}} & \textcolor{blue}{\textit{53.89}} \\
\textit{NPVP} & \checkmark & & \checkmark & \checkmark & \checkmark & \checkmark & \checkmark & \textbf{34.07} & \textbf{0.972} & \textbf{25.59} & \textbf{30.37} & \textbf{0.941} & \textbf{52.18} \\ \hline

\end{tabular}
\caption{Ablation Study on KTH dataset, trained with random contexts. $FL_1$ denotes feature space $L_1$ loss. $PL_1$ denotes pixel space $L_1$ loss. \textit{NPVP} is the stochastic counterpart of model 5. \textbf{Boldface}: best results. \textcolor{blue}{\textit{Blue}}: second best results.}
\label{tab:ablation_study}
\end{table*}

In Figure \ref{fig:OneForAll}, we present examples of one model for all four different conditional video prediction tasks. The first row is the ground-truth (GT) frames. The frames inside a red box are target frames generated by the model given the other context frames. We can observe that the target frames visual quality of VRC and VFI is better than the VPE and VFP, because VPE and VFP only observe the future or past frames, thus there are more uncertainties and it is harder to predict target frames. On the contrary, the context frames of VRC and VFI are scattered across the temporal dimension, which provides more accurate context motion information about the ground-truth event.  Because the model tends to minimize the loss quickly by solving the easier tasks, the model trained with random contexts needs more epochs to reach a comparable performance on VFP and VPE tasks compared to a model trained specifically for that task.

Besides, we investigated the quantitative change of target frames visual quality w.r.t. the varying number of context frames for VFP and VFI with our unified model (Figure \ref{fig:varying_contexts}). The results demonstrate that all three metrics of quality monotonically improves as more context frames are fed into the model. That is, the model generates more accurate target frames given more context frames, which is in alignment with the property of NPs \cite{garnelo2018}. Figure \ref{fig:varying_contexts} also shows the performance gap between VFP and VFI, which indicates that VFI is an easier task. Results of the task-specific \textit{NPVP} (\textit{10$\rightarrow$10}) for VFI and VFP are plotted in Figure \ref{fig:varying_contexts}. The unified model outperforms the task-specific models in terms of all metrics for both VFI and VFP. Therefore, the results validate our motivation that multi-task learning is beneficial.

Finally, a continuous video random missing frames completion (C-VRC) (Figure \ref{fig:OneForAll}, bottom) experiment was conducted to show the continuous generation ability of our model. We observe that it is capable of generating frames at unseen temporal coordinates, such as $4.25, 4.5, 8.5$. Therefore, we can do conditional video prediction at an arbitrary high frame rate. We observe from the generated videos that the continuous predictions by the unified model have a much better temporal consistency than task-specific models. This is because the random contexts help the model to learn more complex conditional distributions \cite{kim2019a}, thus improving the generalization ability of INRs.

\subsection{Ablation study}

We summarize the ablation study results in Table \ref{tab:ablation_study}. The base model (model 1) is a deterministic model (no target event encoding) with a  $\mathcal{T}_D$ with 6 layers ($\mathcal{T}_D$-$6L$), which is trained only by a $L_1$ loss over feature space, and spatio-temporal coordinates are directly fed into the model, i.e., without INR. We gradually modify the architecture and loss function to investigate the influence of the various components. All models are trained with random contexts and thus we evaluate the performance of both VFI and VFP.

\textbf{INR.} We observe a significant performance boost on all metrics in model 2, which validates the effectiveness of INR. Qualitative results also show that predictions of model 1 lack temporal consistency, and have unnatural, choppy motion. \textit{We believe INR improves performance in three ways}: 1) learned Fourier features represent high-frequency details better, contributing to better visual quality. 2) INR learns continuous mapping from coordinates to video pixels, improving continuous prediction. 3) Learnable Fourier features are better multi-dimensional positional encoding, capturing more complex relationships \cite{li2021learnable}.

\textbf{Pixel $L_1$ loss.} By introducing the pixel $L_1$ loss, we observe performance improvement in terms of almost all metrics. Particularly for LPIPS, model 3 outperforms model 2 by a large margin. It proves that pixel $L_1$ loss is beneficial for the target frames visual quality.

\textbf{Size of $\mathcal{T}_D$.} In order to investigate the influence of the size of $\mathcal{T}_D$, we increase the number of layers of $\mathcal{T}_D$ from 6 to 8 ($\mathcal{T}_D$-$8L$). Comparing the results of model 4 with model 3, we observe a performance boost for all metrics of VFI and VFP, which indicates that a larger $\mathcal{T}_D$ is useful. 

\textbf{Context Transformer encoder $\mathcal{T}_E$.} For Model 1-4, there is no explicit temporal relationship learning among context frame features. Because a good aggregation of the context information is critical to improve the performance of NPs, we now include the context Transformer $\mathcal{T}_E$, which models the temporal relationship of $Y_C$ and generates $M_C$ for the $\mathcal{T}_D$ and $z_e$. Comparing model 5 with the previous models, we observe further improvement over all metrics for both VPF and VFI, especially for the LPIPS metric.

\textbf{Stochastic vs Deterministic.} Finally, we modified model 5 to be stochastic by introducing the VAE architecture to give our proposed method \textit{NPVP}. \textit{NPVP} outperforms model 4 for both VFI and VFP in terms of all metrics as we expected, because the event variable takes into account the randomness of prediction instead of only predicting the average of all possible outcomes as its deterministic counterpart \cite{babaeizadeh2018}. We propose a two-stage training strategy to stabilize the training of \textit{NPVP}. For the initial stage, we ignore target event encoding path and $D_{KL}$, in other words, we trained a model 5 first, then the target event encoding path and $D_{KL}$ are included for training.

\section{Conclusion}

We proposed a novel continuous conditional video prediction model based on a neural process and implicit neural representations. By training with random contexts, we can address multiple conditional video prediction tasks with only one model, including future frame prediction, frame interpolation, past frame extrapolation and random missing frame completion. Importantly, all the tasks can be tackled with an arbitrary high frame rate. Results show that our model achieves the SOTA for video future frame prediction and video frame interpolation over multiple datasets.

{\small
\bibliographystyle{ieee_fullname}
\bibliography{references}
}

\clearpage
\begin{center}
\textbf{\large Supplementary Material: A unified model for continuous conditional video prediction}
\end{center}

\setcounter{equation}{0}
\setcounter{figure}{0}
\setcounter{table}{0}
\setcounter{page}{1}
\makeatletter
\renewcommand{\theequation}{S\arabic{equation}}
\renewcommand{\thefigure}{S\arabic{figure}}
\renewcommand{\thetable}{S\arabic{table}}

%
 




\and

\section*{A. Table of important acronyms and notations}
\begin{table}[!h]
\setlength{\tabcolsep}{3pt}
\centering
\begin{tabular}{cc} \hline
NPVP: & \begin{tabular}{@{}c@{}}Neural process \\ for conditional video prediction\end{tabular} \\
\hline
VFI: & Video frame interpolation \\
VFP: & Video future frame prediction \\
VPE: & Video past frame extrapolation \\
VRC: & Video random missing frames completion \\ \hline
NPs: & Neural processes \\
INRs: & Implicit neural representations \\
FFN: & Fourier feature network \\
SIREN: & Sinusoidal representation networks \\
MLP: & Multiple layer perceptron \\
CNN: & Convolutional neural network \\
ConvLSTMs: & Convolutional-LSTMs \\ 
\hline
$V_C$: & Context video frames \\
$V_T$: & Target video frames \\
$X_C$: & Context coordinate representations \\
$Y_C$: & Context video frame features \\
$X_T$: & Target coordinate representations \\
$Y_T$: & Target video frame features \\
$M_C$: & Output feature of $\mathcal{T}_E$ given $X_C$ and $Y_C$ \\
$M_T$: & Output feature of $\mathcal{T}_E$ given $X_T$ and $Y_T$ \\
$z_e$: & event variable \\
\hline
$\mathcal{T}_E$: & Transformer encoder \\
$\mathcal{T}_D$: & Transformer decoder \\
$E_C$: & Context event CNN encoder \\ 
$E_T$: & Target event CNN encoder \\
\hline
\end{tabular}
\caption{Table of important acronyms and notations}
\label{tab:acronym table}
\end{table}

\section*{B. Implementation details}
\section*{B.1 Datasets}

\noindent\textbf{KTH.} KTH dataset includes grayscale videos of 6 different human actions. Following the experimental setup of previous work, we take persons 1-16 as training set, and persons 17-25 as test set. Random horizontal flips and vertical flips are applied to each video clip as data augmentation.

\noindent\textbf{BAIR.} BAIR dataset includes RGB video clips of a robot arm randomly moving over a table with small objects. The training and test sets are defined by the creators of BAIR. Random horizontal flips and vertical flips are applied to each video clip as data augmentation. 

\noindent\textbf{SM-MNIST.} Stochastic Moving MNIST (SM-MNIST) is a synthetic dataset includes videos of two randomly moving MNIST characters within a square region. There is no data augmentation for SM-MNIST during training.

\noindent\textbf{Cityscapes.} Cityscapes dataset includes high-resolution urban traffic videos of many cities. Note that we do not use any annotation provided by Cityscapes, for example, object classes or segmentation masks. Same as previous work, we use the raw video clips from the $"leftImg8bit \_ sequence \_ trainvaltest.zip"$ of Cityscapes. The frames are firstly center-cropped to be square, then we resize the frames to be the resolution of $128\times 128$. There is no data augmentation for the Cityscapes dataset during training.

\noindent\textbf{KITTI.} KITTI dataset includes traffic videos across multiple scenarios, including city, residential, road etc. We follow the experimental setup of previous works \cite{bei2021}, i.e., randomly select 4 sequences from the raw data of KITTI for testing and use the remaining videos for training. The frames are firstly center-cropped and then resize to be the resolution of $128\times 128$. Random horizontal flips and vertical flips are applied to each video clip as data augmentation. 

\subsection*{B.2 Training details}
\textbf{Training of the autoencoder}. For all datasets, the dimension of visual features is set to be $H = 8, W=8, D=512$. For input with a resolution of $64\times 64$, the frame encoder includes 3 downsampling blocks and 2 residual blocks. For input with a resolution of $128\times 128$, the frame encoder includes 4 downsampling blocks and 3 residual blocks. The number of upsampling blocks for the frame decoder equals to the number of downsampling blocks in the corresponding frame encoder. An Adam optimizer with a learning rate of $1e^{-4}$ is used for the training. 

\textbf{Training of the NPs-based predictor}. For all datasets, $\gamma = 0.01$. For BAIR and SM-MNIST, $\beta = 1e^{-6}$. For KTH, $\beta = 1e^{-8}$. The predictors are trained by AdamW, we take a cosine annealing learning rate scheduler with warm restarts \cite{loshchilov2017} at every 150 epochs, the maximum learning rate is $1e^{-4}$ and the minimum learning rate is $1e^{-7}$. Gradient clipping is applied to $\mathcal{T}_E$ and $\mathcal{T}_D$ during training. Please visit \url{https://npvp.github.io} for the code.

\subsection*{B.3 Architecture of VidHRFormer block}
\begin{figure}[!h]
\centering

\includegraphics[clip, trim=8.2cm 8.5cm 6.5cm 7.5cm, width=0.855\linewidth]{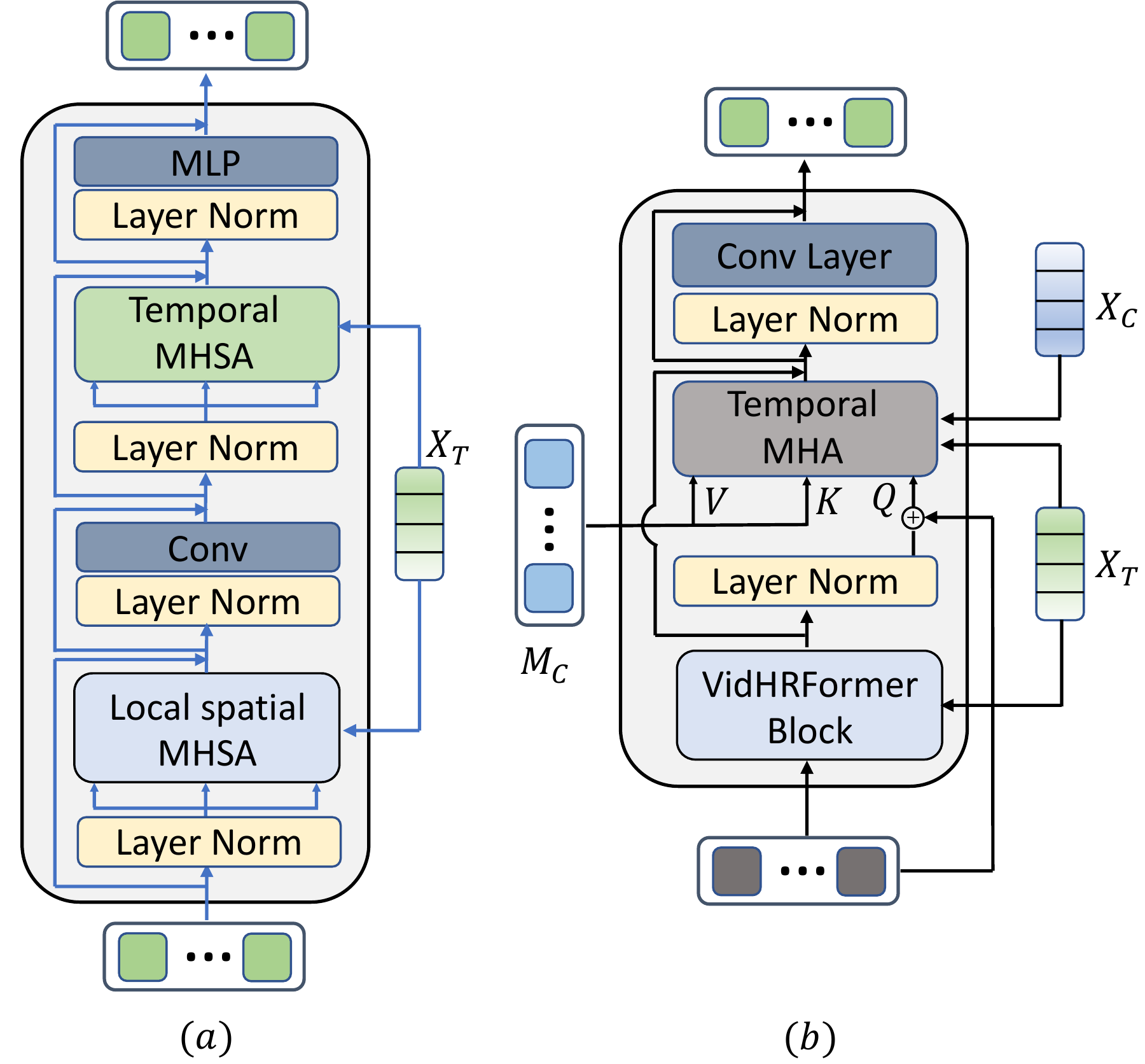}
\caption{(a) VidHRFormer block \cite{ye2022}. (b) Decoder block of VPTR \cite{ye2022}.}
\label{fig:VidHRFormer}
\end{figure}

For the convenience of the readers, we have redrawn the detail architecture of VidHRFormer block \cite{ye2022} and the VPTR decoder block in Figure \ref{fig:VidHRFormer}.

\subsection*{B.4 Architecture of Event encoder $E_C$ and $E_T$}
\begin{figure}[!h]
\centering
\includegraphics[clip, trim=16cm 7.3cm 7.2cm 11cm, width=0.4579\linewidth]{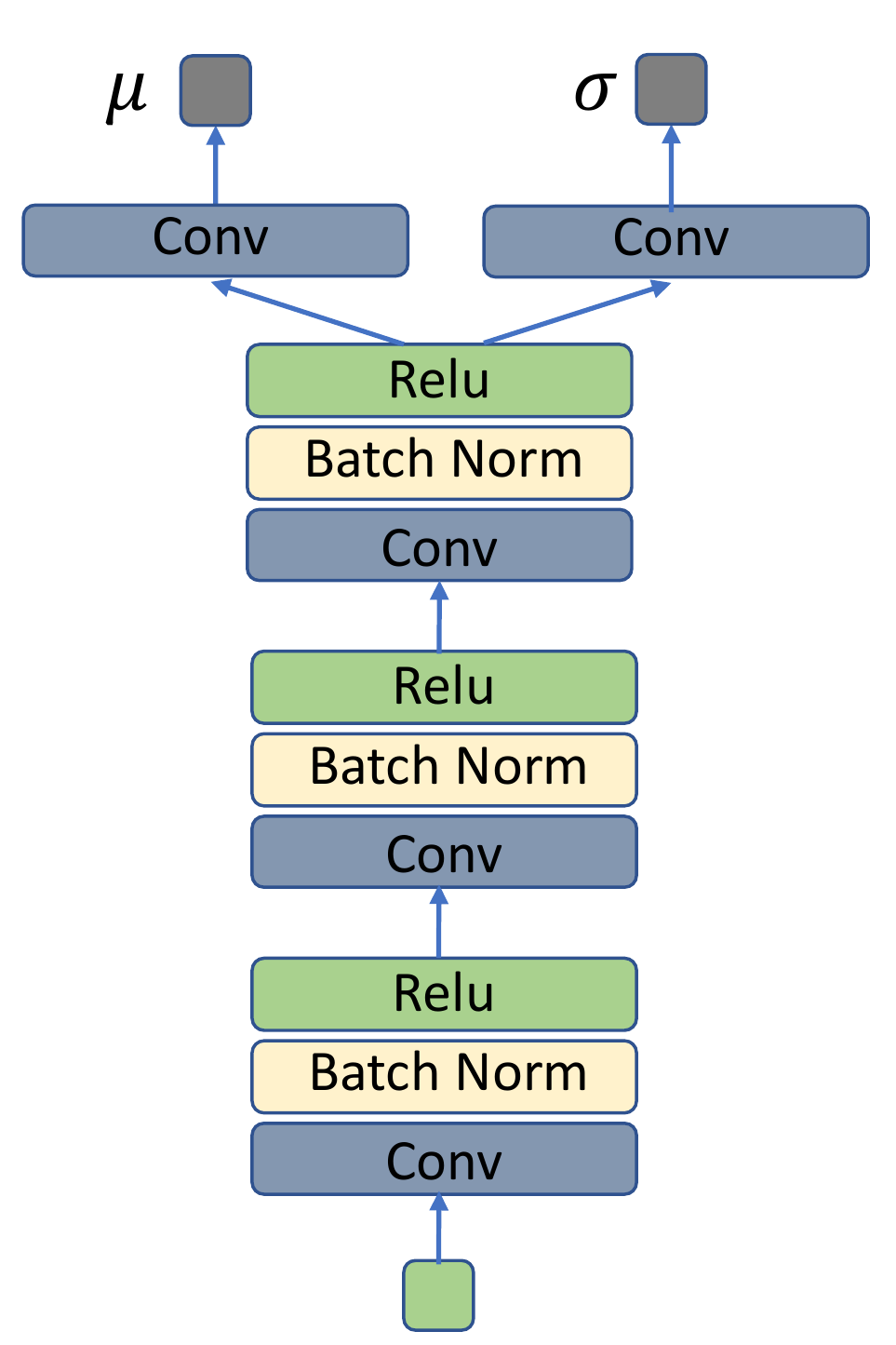}
\caption{Architecture of the Event encoders.}
\label{fig:EventEncoders}
\end{figure}

$E_C$ and $E_T$ share the same architecture, see Figure \ref{fig:EventEncoders}. They are implemented by a small neural network with three $Conv-BN-ReLU$ layers and two $Conv$ heads to output $\mu$ and $\sigma$ respectively.

\section*{C. Qualitative examples}

\subsection*{C.1 Unified model}
\begin{figure*}[!h]
\centering
\includegraphics[clip, trim=5cm 16.5cm 10.1cm 2cm, width=0.8\linewidth]{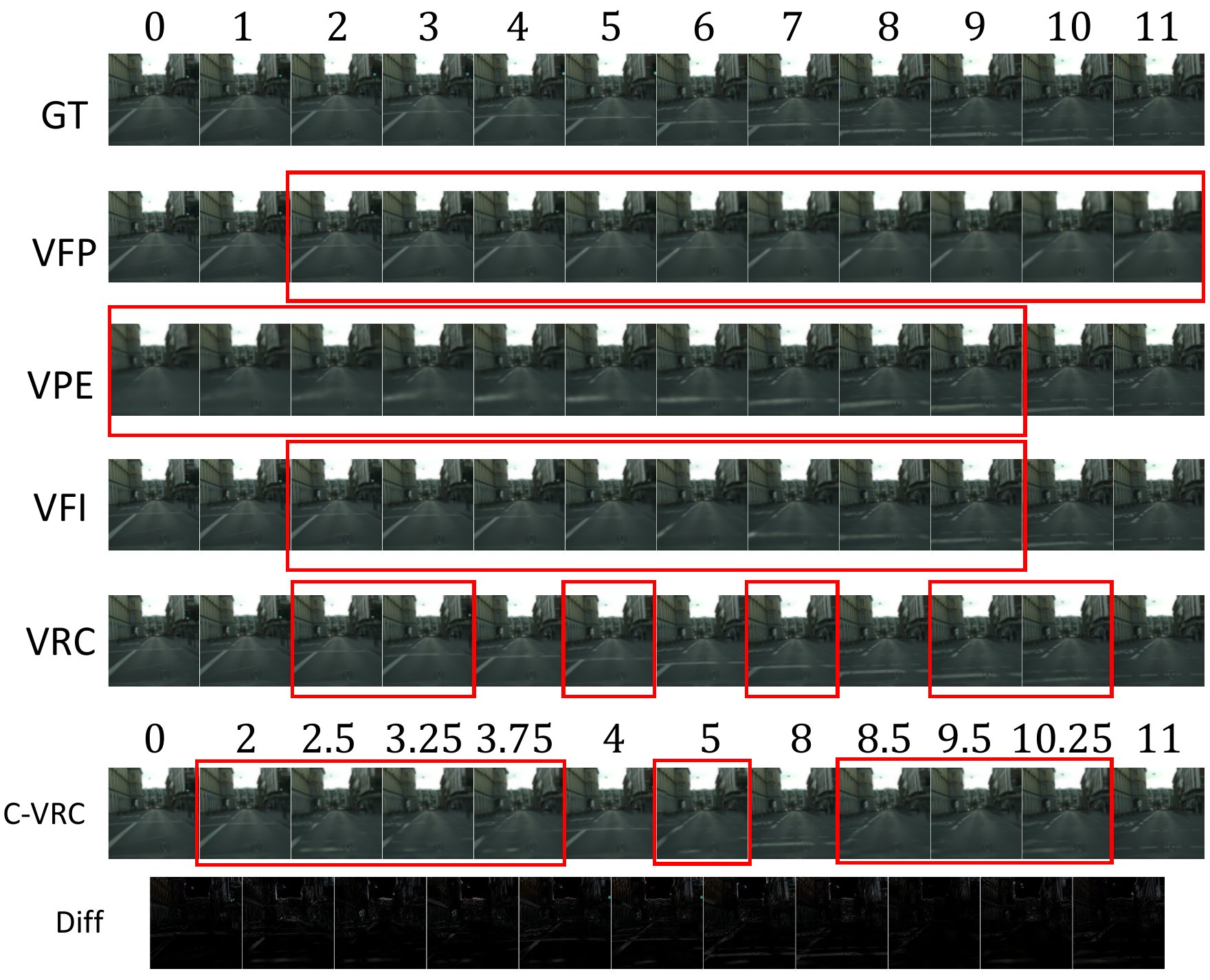}
\caption{One model for all tasks. Frames inside the red boxes are target frames generated by the model. C-VRC denotes continuous VRC. Diff are the difference images between neighboring frames of C-VRC to show that they are all different and that the temporal coordinates are taken into account. }
\label{fig:OneForAllSupp}
\end{figure*}

Here we show another example (see Figure \ref{fig:OneForAllSupp}) of the unified model on Cityscapes dataset for all four different conditional video prediction tasks. In order to demonstrate the continuous prediction ability of NPVP, we take the trained unified model to solve different tasks with different rates, please visit \url{https://npvp.github.io} for video examples of a unified model for KTH dataset.

\subsection*{C.2 Task-specific VFI}
\begin{figure*}[h]
\centering
\includegraphics[clip, trim=1.5cm 6.5cm 10cm 8.5cm, width=0.77\linewidth]{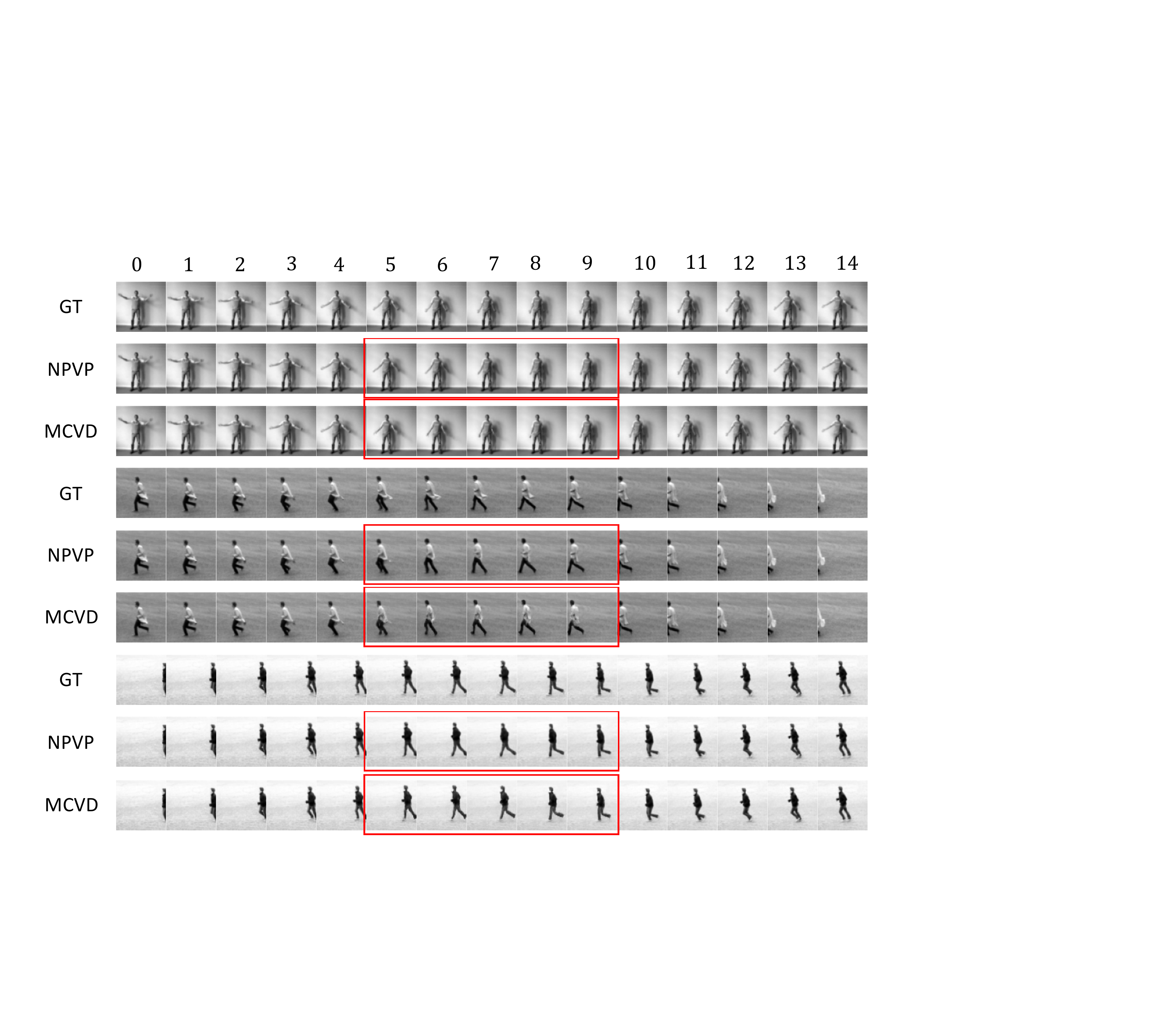}
\caption{VFI examples on KTH by a Task-specific \textit{NPVP} (\textit{10$\rightarrow$5}) model. Frames inside the red boxes are target frames generated by the models. Compared with MCVD \cite{voleti2022}, predicted moving arms or legs by \textit{NPVP} are more realistic and more similar to the ground-truth.}
\label{fig:KTH_specific_VFI}
\end{figure*}

\begin{figure*}[h]
\centering
\includegraphics[clip, trim=1cm 6cm 2cm 2.5cm, width=\linewidth]{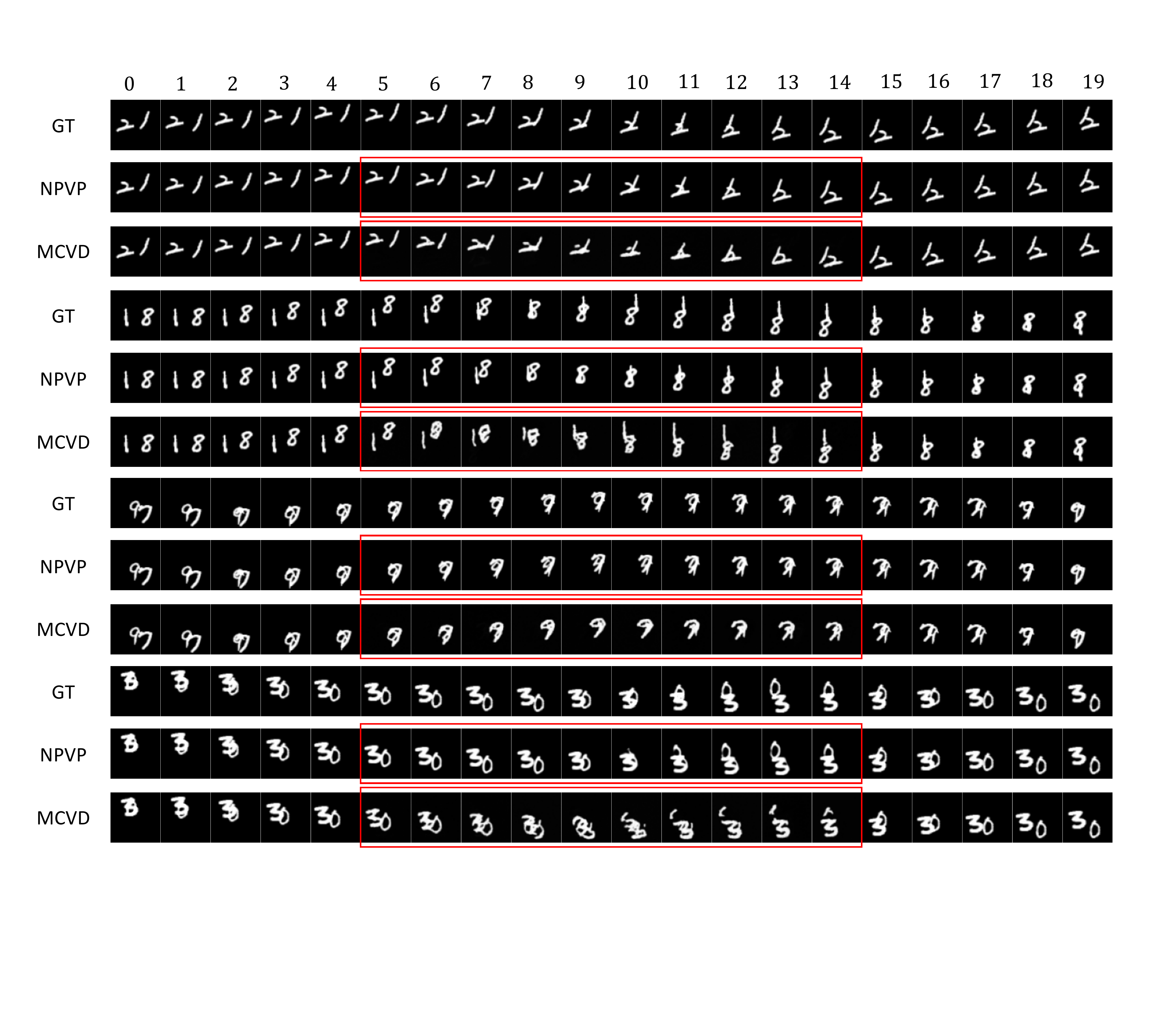}
\caption{VFI examples on SM-MNIST by a Task-specific \textit{NPVP} (\textit{10$\rightarrow$10}) model. Frames inside the red boxes are target frames generated by the model. Compared with MCVD \cite{voleti2022}, the interpolation quality of \textit{NPVP} is better as it captures the shape and motion of MNIST characters for missing frames.}
\label{fig:SMMNIST_specific_VFI}
\end{figure*}

\begin{figure*}[h]
\centering
\includegraphics[clip, trim=1.5cm 12.0cm 1cm 7.5cm, width=\linewidth]{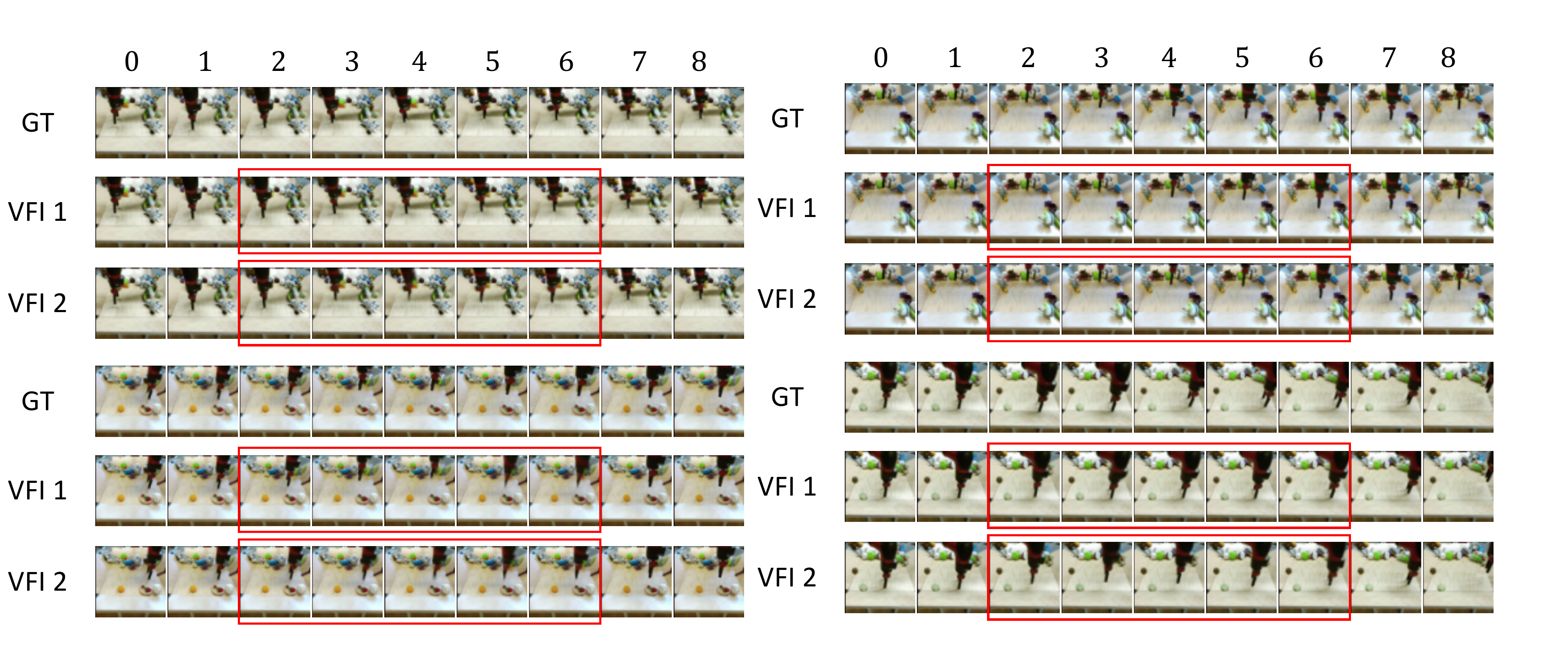}
\caption{VFI examples on BAIR by a Task-specific \textit{NPVP} (\textit{4$\rightarrow$5}) model. Frames inside the red boxes are target frames generated by the model. VFI 1 and VFI 2 denote two different random interpolations given the same contexts.}
\label{fig:BAIR_specific_VFI_rand}
\end{figure*}

We present uncurated VFI examples of KTH, SM-MNIST and BAIR datasets by task-specific \textit{NPVP} models, see Figure \ref{fig:KTH_specific_VFI}, Figure \ref{fig:SMMNIST_specific_VFI} and Figure \ref{fig:BAIR_specific_VFI_rand}. As there is little stochasticity for VFI on KTH and SM-MNIST, we only show the example with the best SSIM from 100 random examples. We also present the VFI results of MCVD \cite{voleti2022} on KTH and SM-MNIST datasets for qualitative comparison. Please visit \url{https://npvp.github.io} for video examples.

\subsection*{C.3 Task-specific VFP}

\begin{figure*}[h]
\centering
\includegraphics[clip, trim=1.5cm 8.0cm 1cm 7.5cm, width=\linewidth]{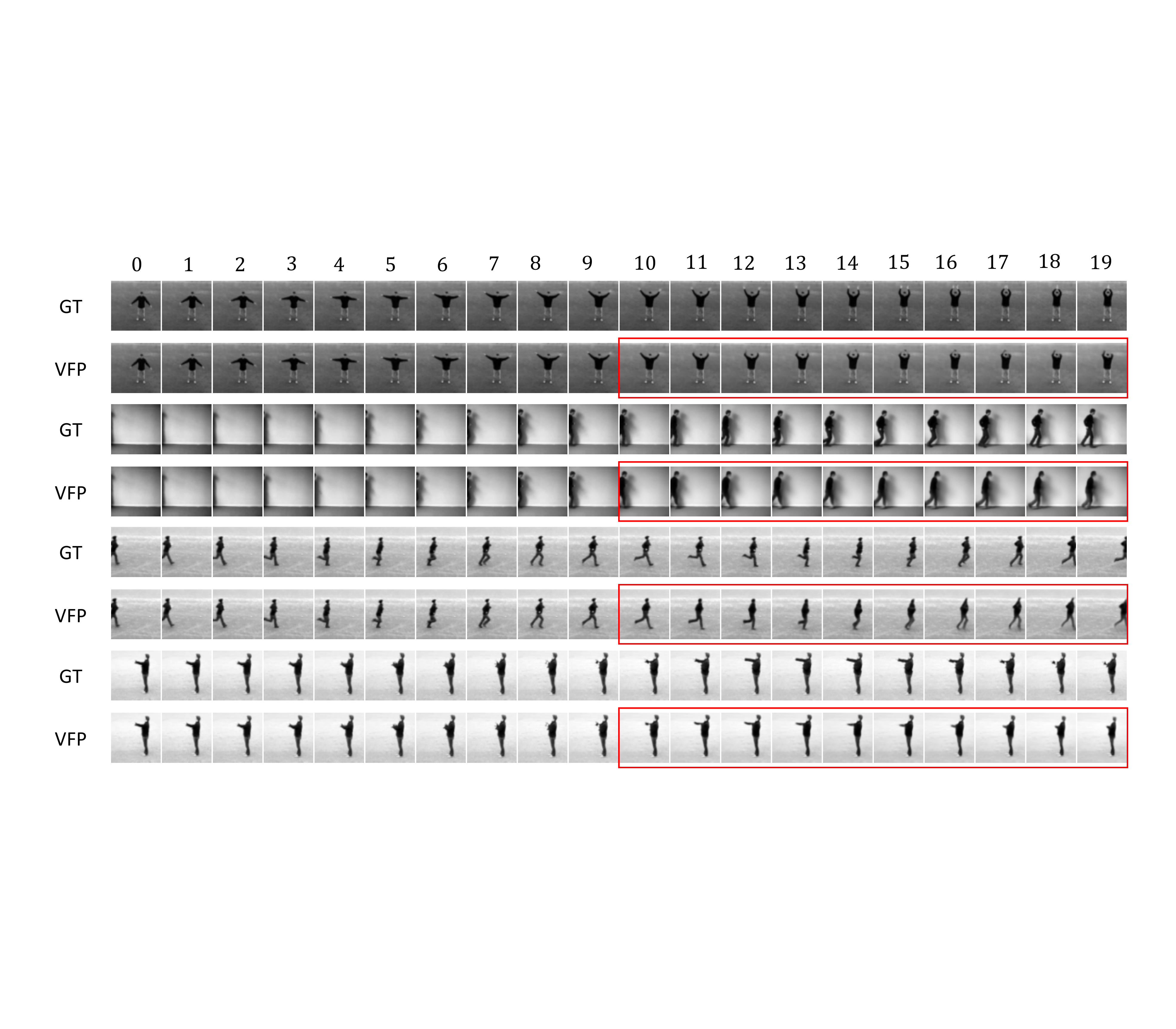}
\caption{VFP examples on KTH by a Task-specific \textit{NPVP} (\textit{10$\rightarrow$10}) model. Frames inside the red boxes are target frames generated by the model.}
\label{fig:KTH_specific_VFP}
\end{figure*}

\begin{figure*}[h]
\centering
\includegraphics[clip, trim=1.5cm 5cm 1cm 5cm, width=\linewidth]{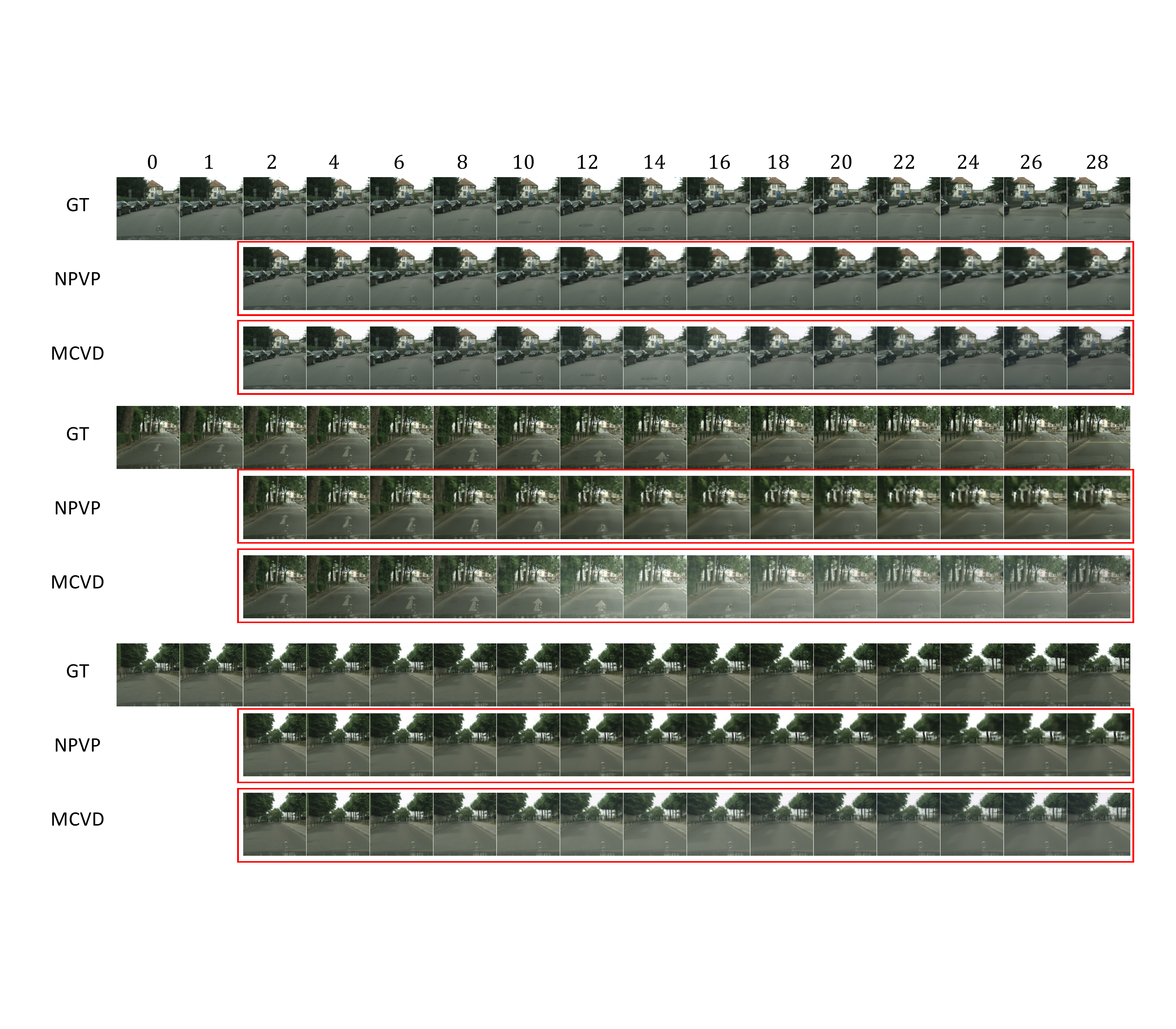}
\caption{VFP examples on Cityscapes by a Task-specific \textit{NPVP} (\textit{2$\rightarrow$28}) model. Frames inside the red boxes are target frames generated by the model. Here we also show the examples generated by MCVD \cite{voleti2022}, which suffers from a brightness-changing problem.}
\label{fig:City_specific_VFP}
\end{figure*}

\begin{figure*}[h]
\centering
\includegraphics[clip, trim=10cm 10cm 8cm 7.5cm, width=0.6\linewidth]{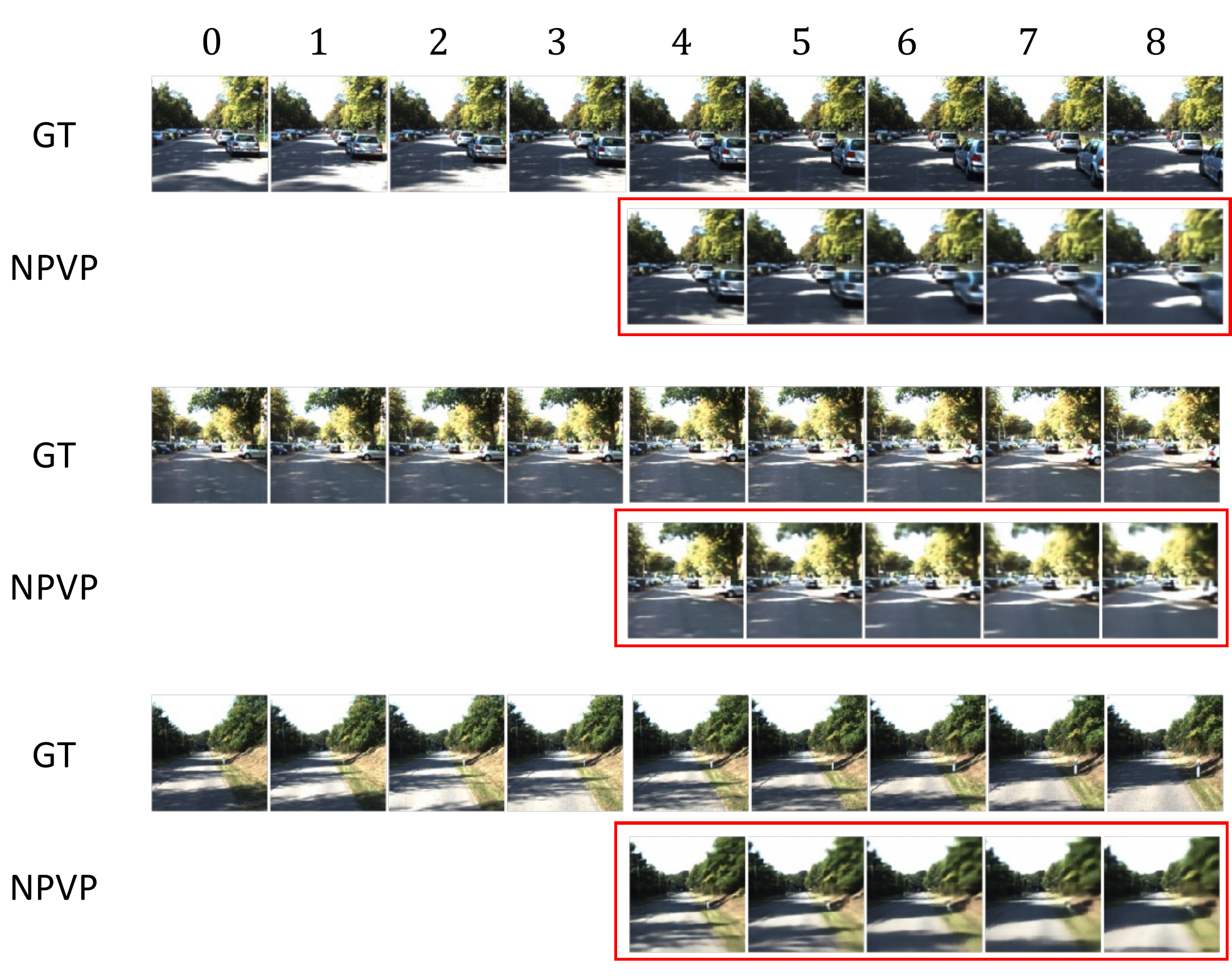}
\caption{VFP examples on KITTI by a Task-specific \textit{NPVP} (\textit{4$\rightarrow$5}) model. Frames inside the red boxes are target frames generated by the model.}
\label{fig:BAIR_specific_VFP_rand}
\end{figure*}

We present VFP examples by task-specific \textit{NPVP} models, see Figure \ref{fig:KTH_specific_VFP}, Figure \ref{fig:City_specific_VFP} and Figure \ref{fig:BAIR_specific_VFP_rand}. For Cityscapes dataset, we show the results of MCVD for comparison. Please visit \url{https://npvp.github.io} for video examples.

\end{document}